\documentclass[pdflatex,sn-mathphys]{sn-jnl}

\usepackage{makecell}

\jyear{2021}%

\theoremstyle{thmstyleone}%
\theoremstyle{thmstyletwo}%

\theoremstyle{thmstylethree}%

\begin{document}

\title[New Challenges in Reinforcement Learning: A Survey of Security and Privacy]{New Challenges in Reinforcement Learning: A Survey of Security and Privacy}


\author[1]{\fnm{Yunjiao} \sur{Lei}}\email{Yunjiao.Lei@student.uts.edu.au}

\author[1]{\fnm{Dayong} \sur{Ye}}\email{Dayong.Ye@uts.edu.au}

\author[1]{\fnm{Sheng} \sur{Shen}}\email{Sheng.Shen-1@student.uts.edu.au}

\author[1]{\fnm{Yulei} \sur{Sui}}\email{Yulei.sui@uts.edu.au}

\author*[1]{\fnm{Tianqing} \sur{Zhu}}\email{Tianqing.Zhu@uts.edu.au}

\author[2]{\fnm{Wanlei} \sur{Zhou}}\email{wlzhou@cityu.edu.mo}

\affil*[1]{\orgdiv{School of Computer Science}, \orgname{University of Technology Sydney}, \orgaddress{\street{Broadway}, \city{Sydney}, \postcode{2007}, \state{NSW}, \country{Australia}}}

\affil*[2]{\orgdiv{School of Data Science}, \orgname{City University of Macau}, \orgaddress{\city{Macau}, \country{China}}}



\abstract{Reinforcement learning (RL) is one of the most important branches of AI. Due to its capacity for self-adaption and decision-making in dynamic environments, reinforcement learning has been widely applied in multiple areas, such as healthcare, data markets, autonomous driving, and robotics. However, some of these applications and systems have been shown to be vulnerable to security or privacy attacks, resulting in unreliable or unstable services. A large number of studies have focused on these security and privacy problems in reinforcement learning. However, few surveys have provided a systematic review and comparison of existing problems and state-of-the-art solutions to keep up with the pace of emerging threats. Accordingly, we herein present such a comprehensive review to explain and summarize the challenges associated with security and privacy in reinforcement learning from a new perspective, namely that of the Markov Decision Process (MDP). In this survey, we first introduce the key concepts related to this area. Next, we cover the security and privacy issues linked to the state, action, environment, and reward function of the MDP process, respectively. We further highlight the special characteristics of security and privacy methodologies related to reinforcement learning. Finally, we discuss the possible future research directions within this area. }



\maketitle

\section{Introduction}\label{sec1}

Reinforcement learning (RL) is one of the most important branches of AI. Due to its strong capacity for self-adaptation, reinforcement learning has been widely applied in multiple areas, including health care~\cite{19ying2020oidpr}, financial markets~\cite{94meng2019reinforcement}, mobile edge computing (MEC)~\cite{113chen2022game,115chen2021deep} and robotics~\cite{96kober2013reinforcement}. Reinforcement learning is considered to be a form of adaptive (or approximate) dynamic programming~\cite{80gosavi2009reinforcement} and has achieved outstanding performance in solving complex sequential decision-making problems. Reinforcement learning's strong performance has led to its implementation and deployment across a broad range of fields in recent years, such as the Internet of things (IoT)~\cite{51lei2020deep}, recommend systems~\cite{52li2010contextual}, healthcare~\cite{50yu2021reinforcement}, robotics~\cite{64levine2016end}, finance~\cite{65deng2016deep}, self-driving cars~\cite{66pan2017virtual}, and smart grids~\cite{67franccois2017contributions}, and so on.
Unlike other machine learning techniques, Reinforcement learning has a strong ability to learn by trial and error in dynamic and complex environments. In particular, it can learn from the environment which has minimum information about the parameters to be learned~\cite{71uprety2020reinforcement}, and can as a method to address optimal problems~\cite{112chen2021rdrl, 114chen2021edge}. 

\begin{figure}[htbp]%
\centering
\includegraphics[width=0.9\textwidth]{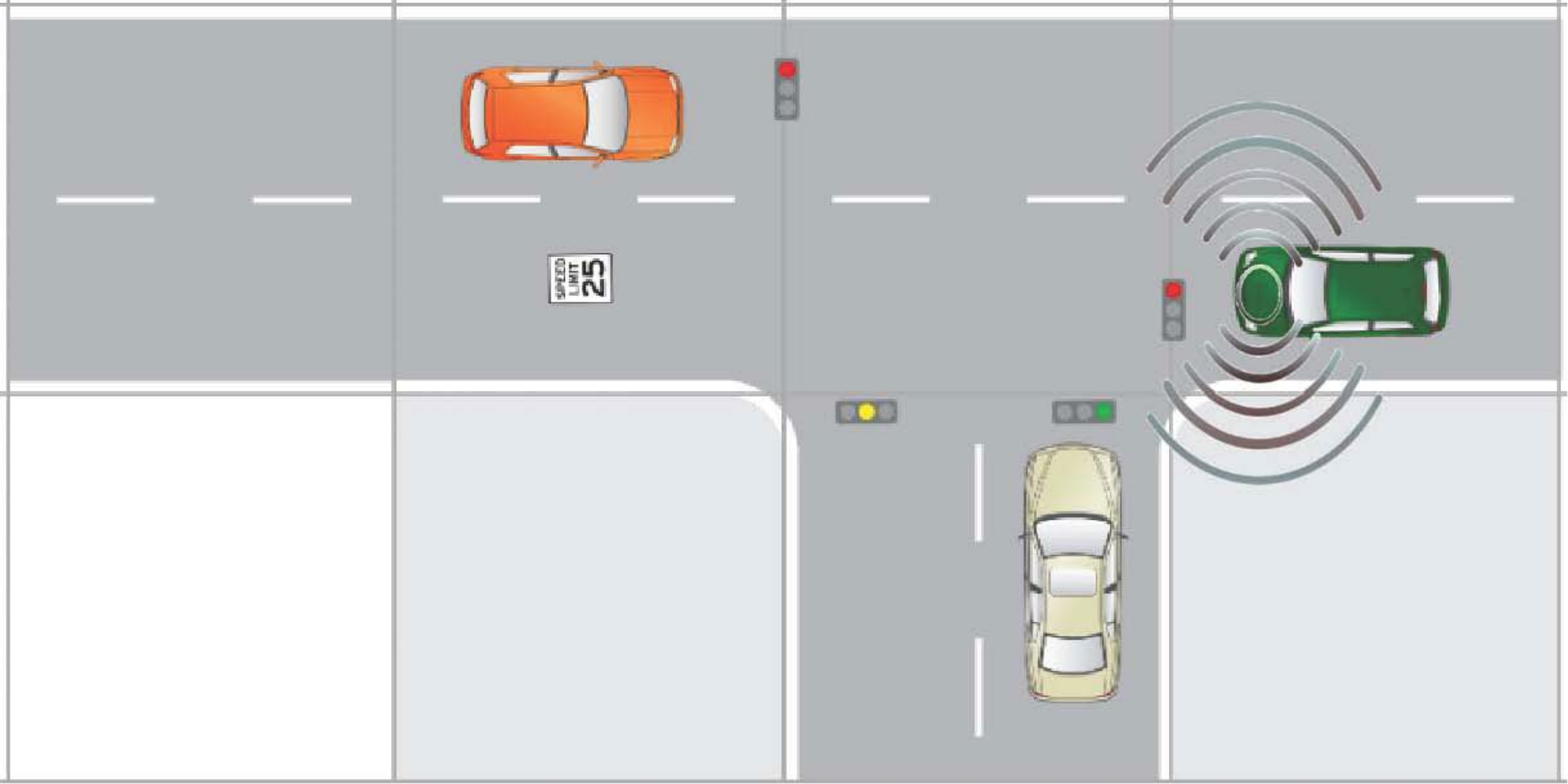}
\caption{An autonomous driving scenario. The green car is an agent. the environment comprises the road, the traffic signs, other cars, etc. }
\label{ev}
\end{figure}

In the reinforcement learning context, an agent can be viewed as a self-contained, concurrently executing thread of control~\cite{92bellifemine2007developing}. It can interact with the environment and obtain a state of the environment as input. The state of the environment can be the situation surrounding the agent's location. Take the road conditions in an autonomous driving scenario as an example. In figure \ref{ev}, the green vehicle is an agent, and all the objects around it can be regarded as the environment; thus, the environment comprises the road, the traffic signs, other cars, etc. Based on the state of the environment, the agent chooses an action as output. Next, the action changes the state of the environment, and the agent will receive a scalar signal that can be regarded as an indicator of the value for the state transition from the environment. This scalar signal is always represented as a reward. The agent's purpose is to learn an optimal policy over time by trial and error in order to gain a maximal accumulated reward as reinforcement. In addition, the combination of deep learning and reinforcement learning further enhances the ability of reinforcement learning~\cite{57mnih2015human}. 

\subsection{Reinforcement learning security and privacy issues }
However, reinforcement learning is weak to security attacks. It is tender for attackers to leverage the breachable data source~\cite{87rakhsha2021policy}. For example, data poisoning attacks~\cite{53huang2019deceptive} and adversarial perturbations~\cite{54behzadan2017vulnerability} are very popular existing approaches in this field. From a defense perspective, several methods have been proposed over the past few years to address these security concerns. Some researchers have focused on protecting the model from attacks and ensuring that the model still performs well while under attack. The aim is to make sure the model takes safe actions that are exactly known, or to get optimal policy under worse situations, such as by using adversarial training~\cite{27wang2020falsification}. 

\begin{figure}[htbp]%
\centering
\includegraphics[width=0.9\textwidth]{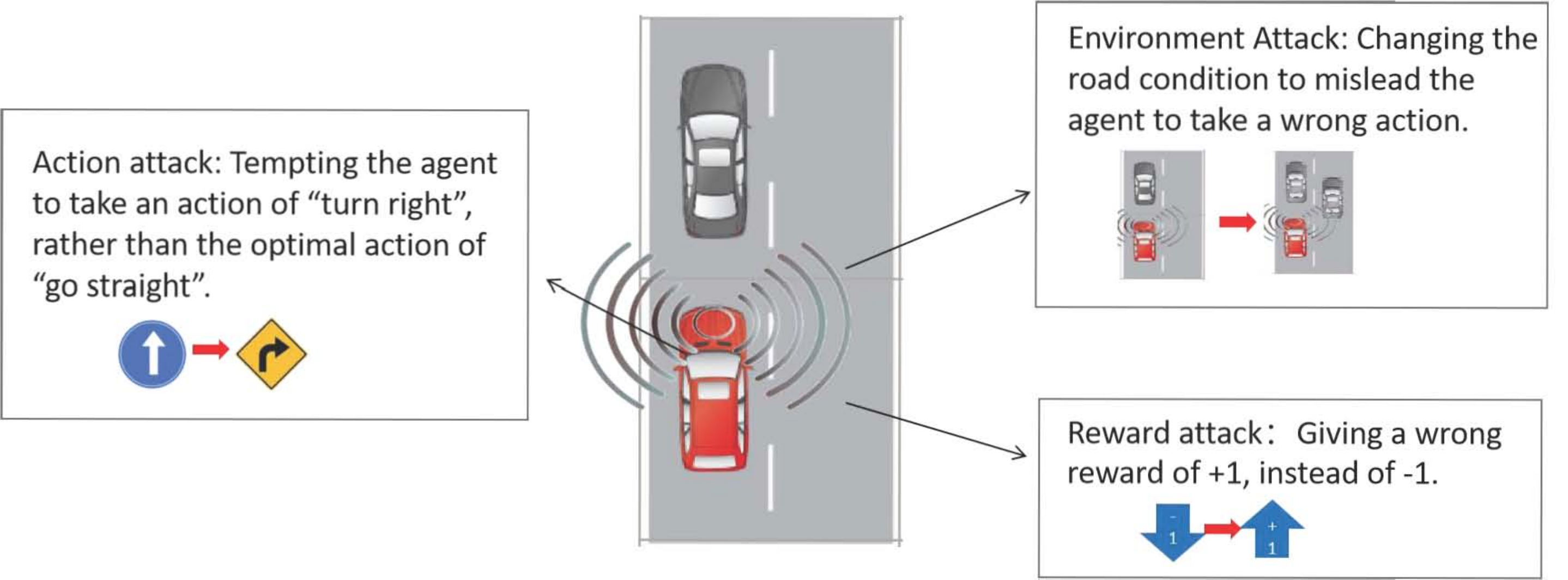}
\caption{A simple example of a security attack in reinforcement learning in the context of automatic driving. An action attack, environmental attack and reward attack are shown respectively. An action attack works by influencing the choice of action directly, such as by tempting the agent to take the action “turn right” rather than the optimal action “go straight”. Environmental attacks attempt to change the agent's perception of the environment so as to mislead it into taking an incorrect action. Finally, the reward attack works by changing the value of a reward given for a specific action in a state. }
\label{e1}
\end{figure}

Figure \ref{e1} presents an example of security attacks in reinforcement learning in an autonomous driving scenario. An autonomous car is driving on the road and observing its environment through sensors. To keep safe while driving autonomously, it will continually adjust its behavior based on the road conditions. In this case, an attacker may focus on influencing the autonomous driving conditions. For example, at a particular time, the optimal action for the car to take is to go straight; however, an action attack may directly influence the agent to turn right(the attack may also impact the value of the reward). With regard to environmental influencing attacks, the attacker may conceive or falsely insert a car in the right front of the environment, and this disturbing may mislead the autonomous car into taking a wrong action. As for reward attacks, rivals may try to change the value of the reward(e.g., from +1 to -1) and thereby impact the policy of the autonomous car. 

Moreover, reinforcement learning also has been subject to privacy attacks due to its weaknesses that can be leveraged by attackers. Established samples used in reinforcement learning contain the learning agent's private information, which is vulnerable to a wide variety of attacks.
For example, in disease treatment applications with reinforcement learning~\cite{19ying2020oidpr}, real-time health data is required, and to achieve an accurate dosage of medicine, the information is always collected and transmitted in plaintext. This may cause disclosure of users' private information; consequently, the reinforcement learning system may collect data from public resources. Most collected datasets contain private or sensitive information that has a high probability of being disclosed~\cite{98zhu2017differentially}. Moreover, reinforcement learning may also require data sharing~\cite{1park2020privacy} and needs to transmit information during the sharing process. Thus, attacks on network links can also be successful in a reinforcement learning context. Furthermore, cloud computing, which is always used for reinforcement learning computation and storage has inherent vulnerabilities to certain attacks~\cite{86xiao2012security}. Rather than changing or affecting the model, the attackers may choose to focus on obtaining or inferring the privacy data; for example, Pan et al.~\cite{6pan2019you} inferred information about the surrounding environment based on the transition matrix.

The main approaches to defending privacy and security in the reinforcement learning context include encryption technology~\cite{4sakuma2008privacy} and information-hiding techniques, such as differential privacy~\cite{59ye2019differentially}.  In addition, some artificial algorithms also have been used to preserve individual privacy~\cite{46sutton2018reinforcement}, such as federated learning (FL) which can preserve privacy for the learning mechanism and structure. Yu et al.~\cite{93yu2020deep} adopt federated learning (FL) into a deep reinforcement learning model in a distributed manner, with the goal of protecting data privacy for edge devices. 

\subsection{Outline and Survey Overview}
As an increasing number of security and privacy issues in reinforcement learning emerge, it is meaningful to analyze and compare existing studies to help spark ideas about how security and privacy might be improved in future in this specific field. Over recent years, several surveys on the security and privacy of reinforcement learning have been completed: 

(1) Chen et al.~\cite{38chen2019adversarial} reviewed the research related to reinforcement learning from the perspective of artificial intelligence security about adversarial attacks and defence. The authors analysed the characteristics of adversarial attack mechanisms and defense technologies respectively.

(2) Luong et al.~\cite{106luong2019applications} presented a literature review on applications of deep reinforcement learning in communications and networking; Such as the Internet of Things (IoT). The authors discussed deep reinforcement learning approaches proposed about issues in communications and networking, which include dynamic network access, data rate control, wireless caching, data offloading, network security, and connectivity preservation. 

(3) Another survey paper~\cite{71uprety2020reinforcement}  conducted a literature review on securing IoT devices using reinforcement learning. This paper presented different types of cyber-attacks against different IoT systems and discussed security solutions based on reinforcement learning against these attacks.

(4) Wu et al.~\cite{72wu2021deep} surveyed the security and privacy risks of the key components of a blockchain from the perspective of machine learning, and help to a better understanding of these methods in the context of IIoT. Chen et al.~\cite{73chen2021deep} also explored deep reinforcement learning in the context of IoT.

Our work differs from the above works.

However, the works mentioned above are all focused on the IoT or communication networks. They are about the application of reinforcement learning. Very few existing surveys have comprehensively presented the security and privacy issues in reinforcement learning rather than the application. Some of them concentrate on the attack and/or defense methods. However, they are just analysing the whole influence. Accordingly, in this paper, we highlight the objects that the attacks aim at and provide a comprehensive review of the key methods used to attack and defend these objects. 

The main contributions of our survey can be summarized as follows: 
\begin{itemize}
\item{The survey organizes the relevant existing studies from a novel angle that is based on the components of the Markov decision process (MDP). We classify current researches on attacks and defences based on their objects in MDP. This provides a new perspective that enables focusing on the target of the methods across the entire learning process.}
\item{The survey provides a clear account of the impact caused by the targeted objects. These objects are components in MDP that are related to each other and may exist in the same time or/and space. Adopting this approach enables us to follow the MDP to comprehend the relevant objects and the relationships between them }
\item{The survey compares the main methods of attacking or defending the components of MDP, and thereby sheds some lights on the advantages and disadvantages of these methods. }
\end{itemize}

The remainder of this paper is structured as follows. We first present preliminary concepts in reinforcement learning systems in Section 2. We then outline the security and privacy challenges in reinforcement learning in Section 3. Next, we present further details on security in reinforcement learning in Section 4, followed by an overview of privacy in reinforcement learning in Section 5. We further discuss the security and privacy in reinforcement learning applications in section 6. Finally, Sections 7 and 8 present our avenues for discussion and future work and conclusion respectively. 

\section{Preliminary}

\subsection{Notation}

Table \ref{table_n} lists the notations used in this article. RL is reinforcement learning, and DRL is deep reinforcement learning. MDP stands for the Markov Decision Process, which is widely used in reinforcement learning. MDP can be denoted by a tuple $(S, A, T, r,\gamma)$, which is made up of the agent action space $A$, the environment state space $S$, the reward function $r$, the transition matrix $T$, and a discount factor $\gamma \in [0,1)$. The transition matrix is a probability mapping from state-action pairs to states $T:(S\times A) \times S\rightarrow[0,1]$. The agent's purpose is to find an optimal policy that can map environment states to agent actions to maximize long-term reward. $v^\pi (s)$ and $Q^\pi (s,a)$ are the state and action-state values, which can regard as a means of evaluating the policy. 

\begin{table}[h]
\begin{center}
\caption{The main notations through the paper. }
\label{table_n}
\begin{tabular}{|c|c|}
\hline
 notations & meaning\\
\hline
$RL$ & Reinforcement learning\\
\hline
$DRL$ &Deep reinforcement learning\\
\hline
$MDP$ & Markov decision process\\
\hline
$A$    & The action space of the agent\\
\hline
$S$   & The state space of the environment\\
\hline
$T$   & The transition matrix\\
\hline
$r$   &  The reward function\\	
\hline
$\gamma$  & A discount factor which is within the range (0,1)\\	
\hline
$ \pi $  & Policy\\	
\hline
$v^\pi (s)$  &  State value\\	
\hline
$Q^\pi (s,a)$ & Action-state value \\	
\hline
\end{tabular}

\end{center}
\end{table}

\subsection{Reinforcement learning}

The reinforcement learning model contains the environment states $S$, the agent actions $A$, and scalar reinforcement signals that can be regarded as rewards $r$. All the elements and the environment can be conceptualized as a whole system. 
At step $t$, when an agent interacts with the environment, it can receive a state of the environment $s_t$ as input. Based on the state of the environment $s_t$, the agent chooses an action $a_t$ using the policy $\pi$ as output. Next, the action changes the state of the environment to $s_{t+1}$. At the same time, the agent will obtain a reward $r_t$ from the environment. This reward is a scalar signal that can be regarded as an indicator of the value for the state transition. 

In this process, the agent learns a piece of knowledge, which may be recorded as $s_t, a_t,r_t,s_{t+1}$ in a Q table. Q table has calculated the maximum expected future rewards for action at each state, and can guide agents to choose the best action at each state. In the next step, the updated $s_{t+1}$ and $r_{t+1}$ will be sent to the agent again. The agent's purpose is to learn an optimal policy $\pi$ so as to gain the highest possible accumulated reward $r$. To arrive at the optimal policy $\pi$, the agent can train by applying a trial and error approach over the long-term episodes. 

A Markov Decision Process (MDP) with delayed rewards is used to handle reinforcement learning problems, such that MDP is a key formalism in reinforcement learning. 
\begin{figure}[htbp]%
\centering
\includegraphics[width=0.9\textwidth]{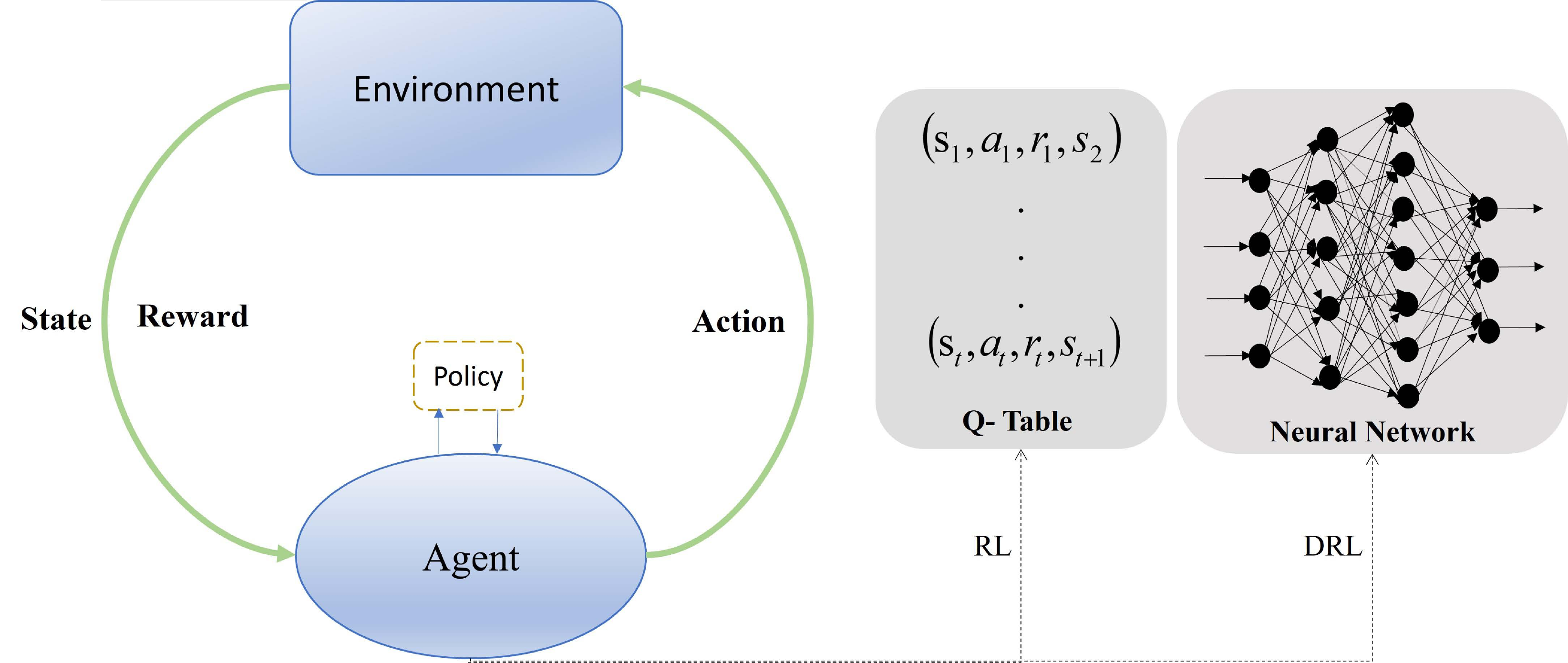}
\caption{The interaction between agent and environment with MDP. The agent interacts with the environment to gain knowledge, which may be recorded as a table or a neural network model (in DRL), and then takes an action that will react to the environment state. } 
\label{RL}
\end{figure}

If the environment model is given, two simple iterative algorithms can be chosen to arrive at an optimal model in the MDP context: namely, value iteration~\cite{55bellman1957dynamic} and policy iteration~\cite{56littman2013complexity}. When the information of the model is not known in advance, the agent needs to learn from the environment to obtain this data based on an appropriate algorithm, which is usually a kind of statistical algorithm. Adaptive Heuristic Critic and $TD(\lambda)$, which is a policy iteration mechanism, were used in the early stages of reinforcement learning to learn an optimal policy with samples from the real world~\cite{47barto1983neuronlike}. Subsequently, the Q-learning algorithm increased in popularity~\cite{48watkins1989learning,49watkins1992q} and is now also a very important algorithm in reinforcement learning. The Q-learning algorithm is also an iterative approach used to select an action with a maximum Q value, which is an evaluation value, in order to ensure that the chosen policy is optimal. Moreover, due to its ability to deal with high-dimensional data and to approximate the function, deep learning has been combined with reinforcement learning to create the field of “deep reinforcement learning” (DRL)~\cite{7sun2020stealthy}. This combination has led to significant achievements in several fields, such as learning from visual perceptual~\cite{57mnih2015human} and robotics~\cite{63gandhi2017learning}.

\begin{figure}[htbp]%
\centering
\includegraphics[width=0.9\textwidth]{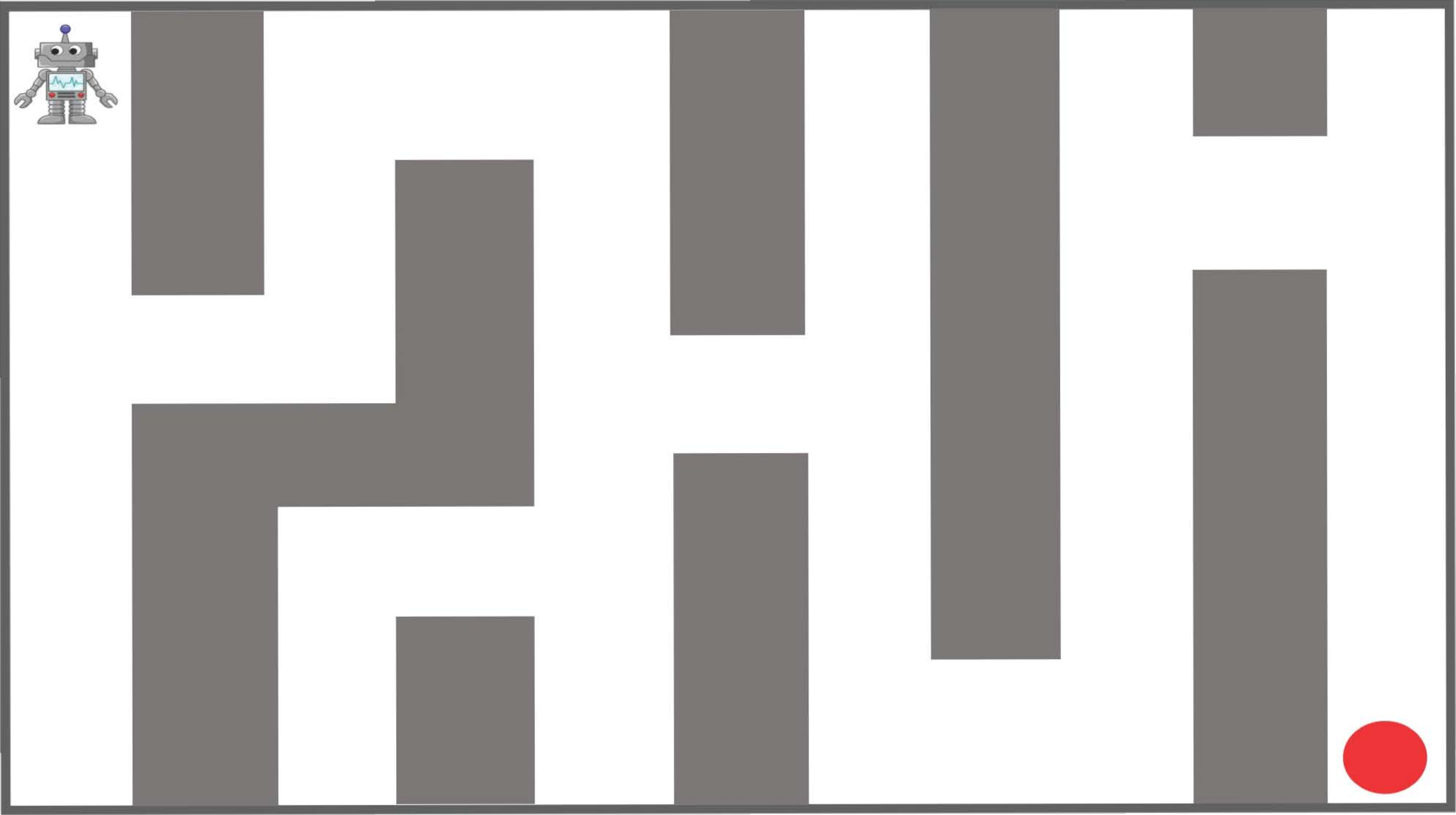}
\caption{A simple example of reinforcement learning, in which a robot tries to find an object in the Grid World environment. The blue robot can be seen as the agent in reinforcement learning. The red circle is the target object. The grey boxes denote the obstacles, while the white boxes denote the road. The robot's purpose is to find a route to the red circle. }
\label{e2}
\end{figure}

An example of reinforcement learning is presented in Figure \ref{e2}. The figure depicts a robot searching for an object in the Grid World environment. The red circle represents the target object, the grey boxes denote the obstacles, and the white boxes denote the road. The robot's purpose is to find a route to the red circle. At each step, the robot has four choices of action: walking up, down, left and right. In the beginning, the agent receives information from the environment which may be obtained through sensors such as radar or camera. The agent then chooses an action and receives a corresponding reward. In the position shown in the figure, choosing the action of up, left or right, may result in a lower reward, as there are obstacles in these three directions. However, taking the action of moving down will result in a higher reward, as it will bring the agent closer to its goal. 

\subsection{Markov Decision Process (MDP)}

The Markov decision process (MDP) is a framework used to model decisions in an environment~\cite{3wang2019privacy}. From the perspective of reinforcement learning, MDP is an approach which has a delayed reward. In MDP, the state transitions are not related to any previous environment states or agent actions. That is to say, the next state is independent of the previous states and based on the current environment state. 

MDP can be denoted as the tuple $(S, A, T, r,\gamma)$, which is made up of the agent action space $A$, the environment state space $S$, the reward function $r$, the transition matrix $T$, and a discount factor $\gamma \in [0,1)$. The transition matrix can be defined as a probability mapping from state-action pairs to states $T:(S\times A) \times S\rightarrow[0,1]$. The agent's purpose is to find an optimal policy $ \pi $ that can map environment states to agent actions in a way that maximizes its long-term reward. The discount factor $\gamma$ is applied to the accumulated reward to discount future rewards. In many cases, the goal of a reinforcement learning algorithm with MDP is to maximize the expected discounted cumulative reward. 

At time step $t$, we denote the environment state, agent action, and reward by $s_t$, $a_t$ and $r_t$ respectively. Moreover, we use $v^\pi (s)$ and $Q^\pi (s,a)$ to evaluate the state and action-state value. The state value function can be expressed as follows:

\begin{equation}
    V^{\pi} (s)=E_{\pi} \left[\sum_{k = 0}^{\infty} {\gamma^{k} r_{t+k+1} \vert s_t=s,\pi} \right]
\end{equation}
The action-state value function is as follows:
\begin{equation}
    Q^\pi (s,a)=E_\pi \left[\sum_{k = 0}^{\infty }{\gamma^{k} r_{t+k+1} \vert s_t=s,a_t=a,\pi} \right]
\end{equation}
where $\gamma$ is the discount factor and $r_{t+k+1}$ is the reward of $t+k+1$ step.
In a wide variety of works, Q-learning was the most popular iteration method applied to discounted infinite-horizon MDPs. 

\subsection{Deep reinforcement learning}

In some cases, reinforcement learning finds it difficult to deal with high-dimensional data, such as visual information. Deep learning enables reinforcement learning to address these problems. Deep learning is a type of machine learning that can use low-dimensional features to represent high-dimensional data through the application of a multi-layer Artificial Neural Network (ANN). Consequently, it can work with high-dimensional data in fields such as image and natural language processing. Moreover, deep reinforcement learning (DRL) combines reinforcement learning with deep neural networks, thereby enabling reinforcement learning to learn from high-dimensional situations. Hence, DRL can learn directly from raw, high-dimensional data, and can accordingly acquire the ability to understand the visual world. Moreover, DRL also has a powerful function approximation capacity, which also employs deep neural networks to train approximate functions in reinforcement learning; for example, to produce the approximate function of action-state value $Q^\pi (s,a)$ and policy $\pi$. 

The process of DRL is nearly the same as that of reinforcement learning. The agent's purpose is also to obtain an optimal policy that can map environment states to agent actions in a way that maximizes long-term reward. The main difference between the DRL and reinforcement learning processes lies in the Q table. As shown in Figure \ref{RL}, in reinforcement learning, this table may be a form that records the map from state to action; by contrast, in deep reinforcement learning, a neural network is typically used to represent the Q table. 

\section{Security and privacy challenges in reinforcement learning}
In this section, we will briefly discuss some representative attacks that cause security and privacy issues in reinforcement learning. In more detail, we explore different types of security attacks (specifically, adversarial and poisoning attacks) and privacy attacks (specifically, genetic algorithm (GA) and inverse reinforcement learning (IRL)). Moreover, some representative defence methods will also be discussed (specifically, differential privacy, cryptography, and adversarial learning). We further present the taxonomy based on the components of MDP in this section, along with the relationships and impacts among these components in reinforcement learning. 

\subsection{Attack methodology} 

\subsubsection{Security attacks}
In this part, we discuss security attacks designed to influence or even destroy the reinforcement learning model in the reinforcement learning context. Specifically, we briefly introduce some recently proposed attack methods developed for this purpose.

One of the popular meanings of the term "security attack" is an adversarial attack with adversarial examples~\cite{10rakhsha2020policy,26lin2020robustness}. The common form of adversarial examples involves adding imperceptible perturbations to data with a pre-defined goal; these perturbations can deceive the system into making mistakes that cause malfunctions, or prevent it from making optimal decisions. Because reinforcement learning gathers examples dynamically throughout the training process, attackers can directly add imperceptible perturbations to states, environment information, and rewards, all of which may influence the agent during reinforcement learning training. For example, consider the addition of tiny perturbations to state $s$ in order to produce $s+\delta$~\cite{25zhao2020blackbox,7sun2020stealthy} ($\delta$ is the added perturbation). Even this small change may affect the following reinforcement learning process. Attackers determine where and when to add perturbations, and what perturbations to add, in order to maximize the effectiveness of their attack. 

Many algorithms that add adversarial perturbations have been proposed. Examples include the fast gradient sign method (FGSM), which can calculate adversarial examples, the strategically-timed attack, which focuses on selecting the time step of adversarial attacks, and enchanting attack (EA), which can mislead the agent regarding the expected state through a series of crafted adversarial examples. Moreover, defenses to adversarial examples have also been studied. The most representative method is adversarial training~\cite{12li2019robust}, which trains agents under adversarial examples and thereby improves model robustness. Other defensive methods focus on modifying the objective function, such as by adding terms to the function or adopting a dynamic activation function. 

Another common type of security attack is the poisoning attack, which focuses on manipulating the performance of a model by inserting maliciously crafted "poison data" into the training examples. A poisoning attack is often selected when an attacker has no ability to modify the training data itself; instead, the attacker adds examples to the training set, and those examples can also work at test time. Attacks based on a poisoned training set aim to influence the behaviour of the model so that it outputs incorrect results. As reinforcement learning requires a very large amount of data for training, and may also collect various types of data from sensors and public applications, it may be vulnerable to fake data injected by attackers into these kinds of data inputs. Examples include the poisoning attack on the environment~\cite{87rakhsha2021policy}, in which the attacker crafts malicious environmental examples(such as the transition matrix) at training time to change the policy. 

The most effective method of crafting poisoned data may be the traditional gradient-based method, and there are also some methods based on the gradient method. The representative methods to defend against poisoning attacks are detection methods and training-based defences. The detection method attempts to detect the poisoned training data or identify corrupted models after they have been trained. The training-based defences are designed to develop robust training routines or to remove the effects of poisoned data. 

\subsubsection{Privacy attacks}

Two common types of privacy attacks are those that get/search private information directly and those that infer private information based on known information. Data transfer and storage are necessary components of reinforcement learning, as the agent requires a large amount of data for training and also needs to interact with its environment. As a consequence, privacy attacks on storage and transfer can be also used for reinforcement learning. Moreover, some other special inferring methods are also used.  

Genetic algorithm (GA) belong to the category of evolutionary computing algorithms, which are a type of search algorithm inspired by the process of natural selection. These algorithms can be used to attack reinforcement learning systems in order to obtain privacy information. Transition matrix search~\cite{6pan2019you} is one such method. The basic Genetic Algorithm starts with a randomly initialized population of candidates. A selection operator selects parents by randomly picking two candidates and choosing the one with the higher score. Crossover is then used to generate the child candidates of selected parents, and random mutation is applied to the child candidates to generate new candidates~\cite{88sehgal2019deep}. 

Inverse reinforcement learning (IRL) is a kind of inferring algorithm that can be used to infer the reward function of reinforcement learning based on the policy or observed behaviour~\cite{89arora2021survey}. In inverse reinforcement learning methods, the observed agent is regarded as an expert while the subject agent is viewed as the learner, and IRL assumes that the expert is behaving according to an underlying policy. The purpose of IRL is to learn an optimal reward function that can explain the observed behaviours. IRL is usually used to help the reinforcement learning system to obtain a reward function; however, because of its ability to infer, IRL may also be applied to attack. 

\subsection{Possible defense methodologies}

In this section, we will present three representative defensive methods that are widely used in various fields. 

\subsubsection{Differential Privacy}
Differential privacy is a prevalent privacy model that can guarantee minimal impact on the analytical output of a dataset if any individual record is stored in or removed from a dataset~\cite{90ye2020differentially}. 

In differential privacy, two datasets $D$ and $D^{'}$ are regarded as neighbouring datasets if they differ in terms of only one record. A query $f$ is a function that maps the dataset $D$ to an abstract range $R$ ($f:D \rightarrow R$). A group of queries is denoted as $F=\lbrace f_1, ... , f_m \rbrace$, and
$F(D)$ denotes $\lbrace f_1(D), ... ,f_m(D)\rbrace$. The maximal 
difference in the results of query $f$ is defined as the sensitivity of query $\Delta f$, which determines how much perturbation is required for a privacy-preserving answer at a given privacy level. The goal of differential privacy is to mask the differences in the answers to query $f$ between the neighbouring datasets. To achieve this goal, differential privacy provides a mechanism $M$, which is a randomized algorithm that accesses the datasets. 

There are three types of widely used differential privacy mechanisms: the Laplace mechanism, the exponential mechanism and the Gaussian mechanism. The Laplace mechanism adds Laplace noise to the true answer; here, Lap(b) is used to represent the noise sampled from the Laplace distribution with scaling $b$. Exponential mechanisms $M$ select and output an element with probability proportionality. Compared to a Laplace mechanism, a Gaussian mechanism adds zero-mean isotropic Gaussian distribution sampled noise. 

Differential privacy can be used in learning problems to improve various aspects of a model, such as randomization, privacy preservation capability, and algorithm stability~\cite{97zhu2020more}. For example, Ye et al.~\cite{100ye2022one} applied differential privacy to a confidence score vector containing a probability distribution over the possible classes predicted by an ML model. This approach can defend against data inference attacks in a time-efficient manner by controlling the utility loss of confidence score vectors; in so doing, it fully reflects the advantages of the privacy preservation capability and algorithm-stability ability of differential privacy.

There are two popular variants of differential privacy: joint differential privacy (JDP) and local differential privacy (LDP). The former has a centralized agent that is responsible for protecting users’ sensitive data, while in the latter, information needs to be protected directly on the user side.

\subsubsection{Cryptography} 
Cryptography is the classic method used for privacy protection fields by encoding messages so that they cannot be understood by untrusted parties. The main encoding techniques are symmetric algorithms and asymmetric algorithms. Symmetric algorithms utilize the same key for both encryption and decryption. Examples include Data Encryption Standard (DES), triple-DES (TDES) and Advanced Encryption Standard (AES). DES was the first encryption standard method. It employs a block cipher that can encrypt 64 bits of plain text at a time, along with a 56-bit key. The TDES algorithm adopts three rounds of DES encryption, so that it has a key length of 168 (56 * 3) bits, along with two or three 56-bit keys. This method first uses three different keys to generate the cipher text $C(t)$ from the plaintext message $t$. One of the most popular Cryptography methods in recent years is Homomorphic Cryptosystems~\cite{4sakuma2008privacy}, which allows operations on the cipher text; hence, it has great adaptability and is suitable for use in different systems for different aims~\cite{91alaya2020homomorphic}. A homomorphic encryption algorithm $H$ is a set of four functions $H = {Key Generation, Encryption, Decryption, Evaluation}$. Here, key generation is a client that generates a pair of keys: a public key $pk$ and a secret key $sk$ for encryption of plain text. The purpose of Encryption is to use the $sk$ client to encrypt the plain text PT and generate Esk(PT). The cipher text (CT) will be sent to the server with the public key $pk$. The Evaluation function evaluates the cipher text (CT). Finally, the Decryption function uses $sk$ to decrypt and obtains the original result. 

\subsubsection{Adversarial learning}
Some smart methods have been adopted for privacy protection purposes. One of these methods, adversarial learning, is used specifically to combat adversarial attacks. The main idea behind adversarial learning involves training a model on a training set with adversarial examples to increase model robustness. To do this effectively, it may be necessary to generate a large number of adversarial examples or increase the amount of perturbed data. During training time, an agent may learn with a modified objective function that has the original loss function $J$. Training with an adversarial objective function based on the fast gradient sign method is one of the popular method. The modified objective function is as follows:
\begin{equation}
  \tilde{J} \left(\theta,x,y \right)=\alpha J \left(\theta,x,y \right)+(1-\alpha) J \left(\theta,x+\epsilon sign(\nabla_x J(\theta,x,y)),y \right)
\end{equation}
where $J$ is the original loss function; $\theta$ is the parameters of a model; $x$ is the input to the model; $y$ is the targets associated with $x$; $\epsilon$ is a small constant that bounds the magnitude of the perturbations $\eta=\epsilon sign(\nabla_x J(\theta,x,y)), \Vert \eta \Vert_{\infty} < \epsilon$; $\alpha$ is also a constant that controls the weighting of the loss terms between normal and adversarial inputs.

Notably, these methods may require a huge number of adversarial examples, and may also be computationally intensive. Moreover, some of the defensive strategies do not work against certain kinds of adversarial attacks.

\subsection{Common attack models in the security and privacy in reinforcement learning}
Reinforcement learning is the same as other machine learning models, we can consider the three main domains in security and privacy problems: White-Box, Black-Box, and Grey-Box models.

A White-Box models' parameters, structures and training methods are transparent. Therefore, the inner logic and decision-making process are interpretable. 

In contrast to the White-Box models, the parameters, structures and training methods of Black-Box models are not known and are hard to interpret. That is to say, the attackers could only know the expected inputs and the corresponding outputs of a black-box model. For example, Zhao et al.~\cite{25zhao2020blackbox} studied adversarial sample attacks in reinforcement learning where the attacker has no knowledge of the reinforcement learning agents, both their training parameters and their training methods. The attackers only can observe the agent playing the game to build up a collection of the agent's observation $s_{t}$, previous actions $A_{t-1}$ and previous states $S_{t-1}$.

A Grey-Box is a combination of Black-Box and White-Box models~\cite{105bohlin2006practical}. Grey-box models assume the attackers have access to some of the agent’s internal parameters, structures and training methods. For example, states or training methods of reinforcement learning agents, or the attackers can access partial information of the target agent or its training environment.

\subsection{Introduction of experiment scenarios for reinforcement learning}

\begin{figure}[htbp]%
\centering
\includegraphics[width=0.75\textwidth]{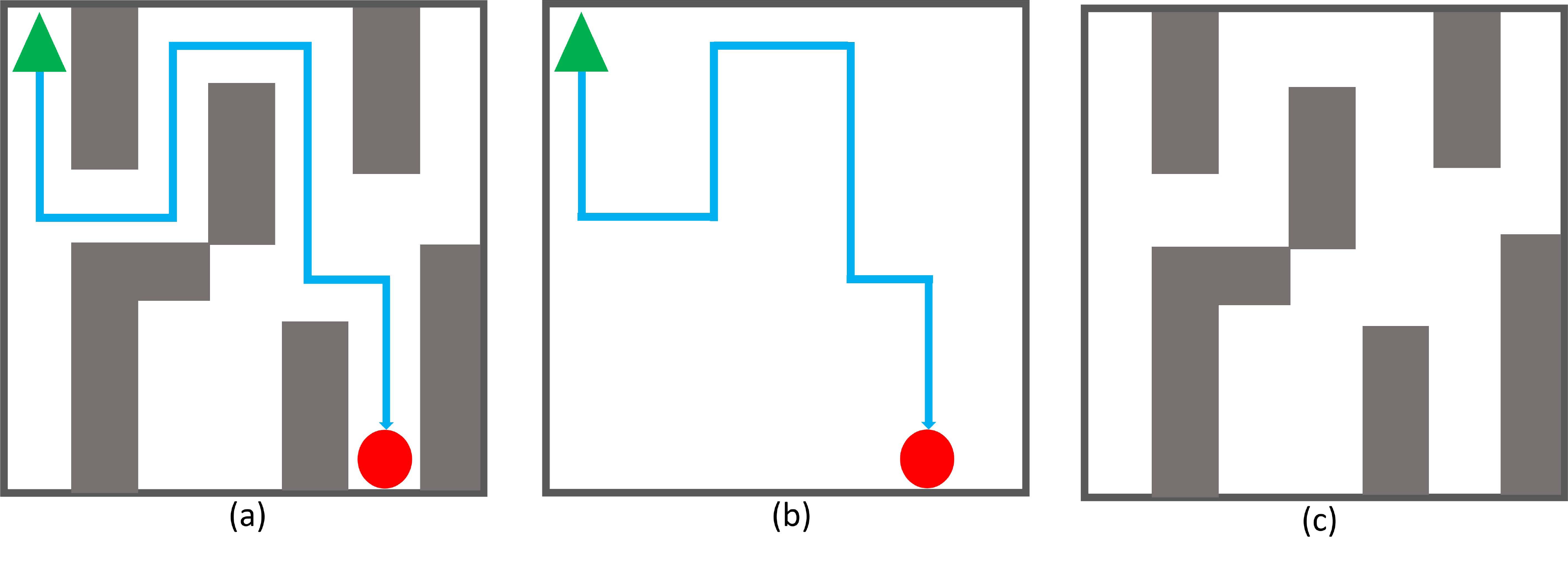}
\caption{Grid world experiment environment. There were obstacles (the grey blocks) and the goal of the agent (the red circle). The agents are marked as a green triangle. The agents' task is to collect the rubbish around the environment. } 
\label{gw}
\end{figure}

There are some popular scenarios for reinforcement learning experiments which can help analyse the methods. Grid World is one of the experiment environments. Grid world is widely used for it is simple and easy to use. It is suitable for most methods. An example of a grid world is shown in Fig.~\ref{gw}. There were grey blocks representing obstacles and a red circle representing the goal of the agent. The agents are marked as a green triangle. We can adjust the scale, the details, and the rewards of the environment based on the specific algorithm. It is easy to the extent of a lot of different methods such as in paper~\cite{8zhang2020adaptive}. The environment in this scenario is the grid world, the agent can take actions of up, down, right, and left. There can be a leakage the private information within the Grid World. For example, a trained reinforcement learning agent based on the policy and towards the goal as shown in Fig.~\ref{gw} (a) (follows the trajectory of the blue line). An attacker can based on the trajectory as the blue line in Fig.~\ref{gw} (b) infer the private information of the environment (as shown in Fig.~\ref{gw} (c)). 

\begin{figure}[htbp]
\centering
\includegraphics[width=0.75\textwidth]{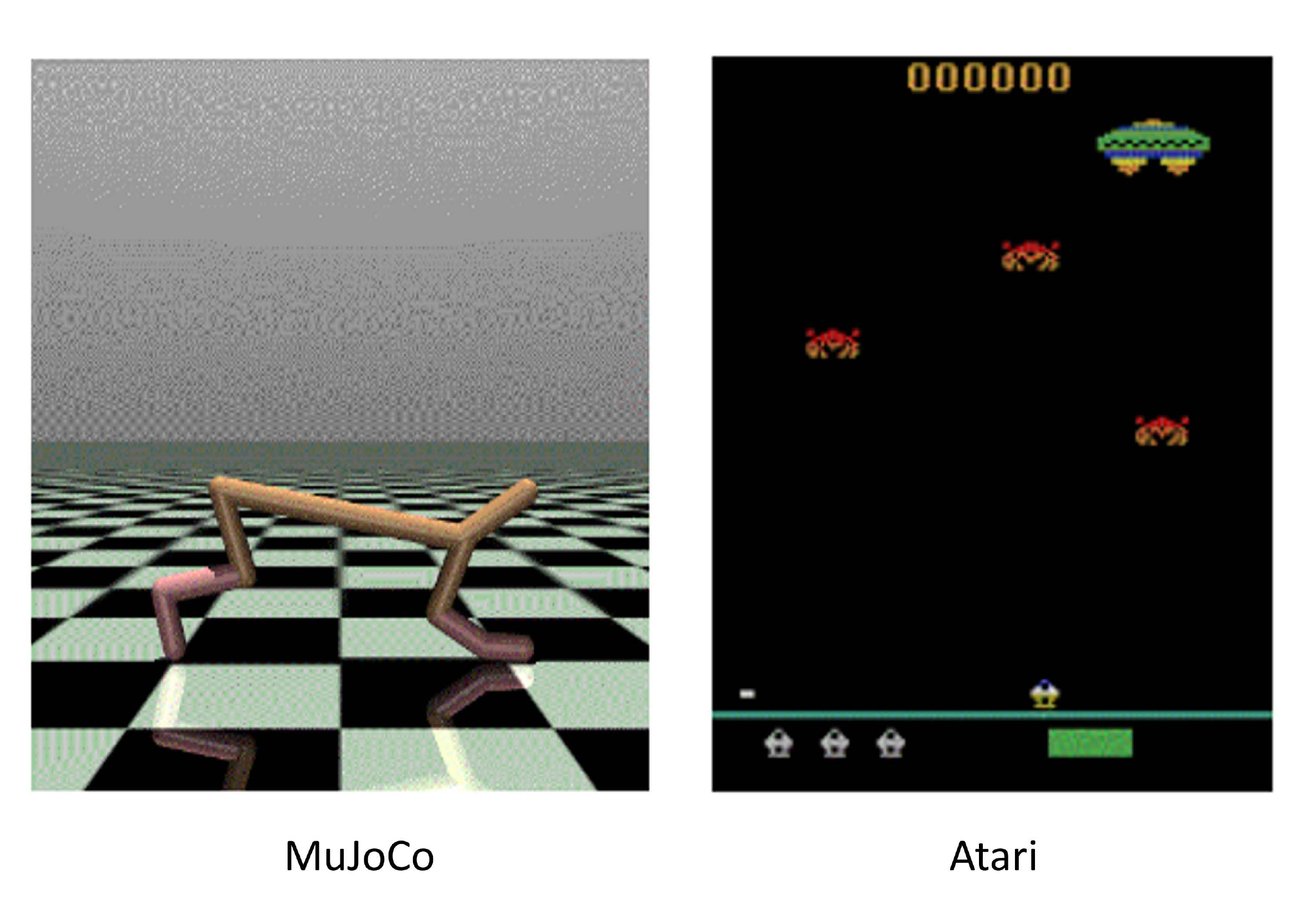}
\caption{Classical Atari games and MuJoCo experiment environment from Open AI resources} 
\label{OAI}
\end{figure}

OpenAI’s gym is an open-source experiment environment which has many different and useful environments. for example, Atari~\cite{118kaiser2019model} and MuJoCo~\cite{14tessler2019action}. The Atari games and MuJoCo environment is like shown in Fig.~\ref{OAI}. For MuJoCo domains Half Cheetah, the agent is learning to apply torque on the joints to make the cheetah run forward (right) as fast as possible. The reward can be set as a positive value when the agent moves forward and a negative value for the agent moves backwards. An action represents the torques applied between links and the state consisting of positional values of different body parts of the cheetah. In the classical Atari games, the goal of the agent is to destroy these enemies and dodge their attacks. An agent can take actions such as noop, fire, up, right, left, rightfire, and leftfire. The state can be the RGB image of the environment. An attacker can affect the states, actions, rewards and so on to influence the performance of the agent.

\begin{figure}[htbp]%
\centering
\includegraphics[width=0.7\textwidth]{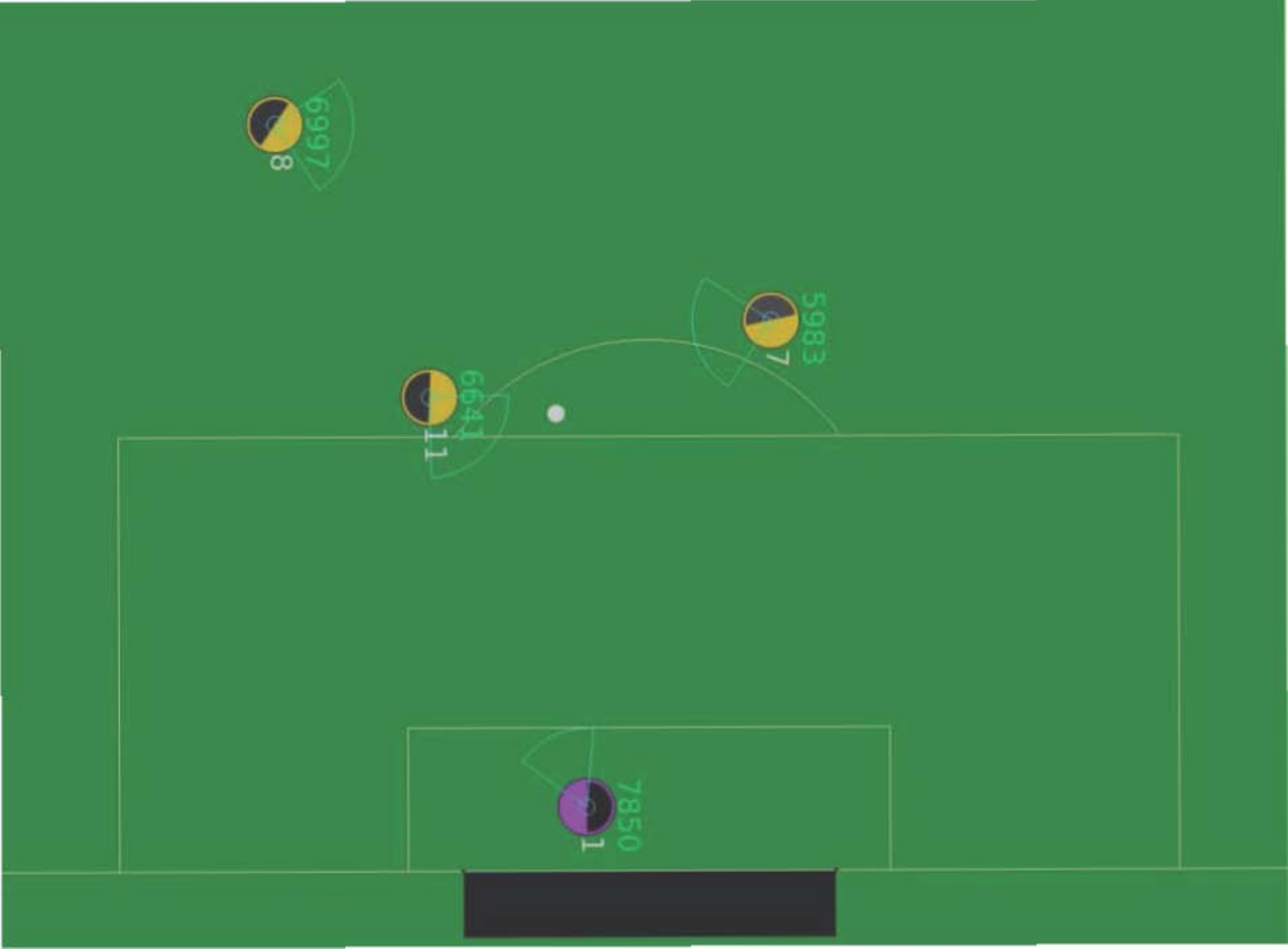}
\caption{ HFO game: three players and one goalkeeper~\cite{99ye2022model}} 
\label{FT}
\end{figure}

For multi-agent reinforcement learning, one of the popular experiment environments is Half Field Offense in Robocup 2D Soccer as shown in Fig.\ref{FT}, which has three players trying to score goals against a goalkeeper. The agents communicate with each other to get the goal. They can take actions like move, shoot, pass, dribble, catch, and noop. They can have information on environment states like goal angle, positions of every agent, the distance between agents and so on. Attackers also can focus on the elements of MDP to influence the cooperation of the agent team.

All the scenarios mentioned above are common experiment environments that can be used for reinforcement learning. If there is no requirement for a special application environment, we can choose them for helping analyse and compare the security and privacy in reinforcement learning issues. We can see that the agent will take actions based on the environment state, and the taken action will decide the received reward and react to the environment. Thus, reinforcement learning is a process, and every element in this process will affect the process chain.  
 
\subsection{Taxonomy based on the components of MDP}
In the below, we classify security and privacy in reinforcement learning based on the components of MDP. First, we categorize the existing research as either security-related or privacy-related. In this paper, we regard papers about attacks that influence or even destroy the reinforcement learning model and defenses that enhance the robustness of the reinforcement learning model as examples of security problems in reinforcement learning. For example, studying how to attack a system to mislead the agent, or how to protect the system from adversarial attacks so that it remains stable and produces good output. In addition, research into obtaining, inferring, or conversely protecting private information will be regarded as related to privacy in reinforcement learning in this paper. Examples include inferring the information of the environment based on known transition matrix and using cryptograms to preserve privacy.

We subsequently conduct a further classification of security in reinforcement learning and privacy in reinforcement learning based on the MDP perspective. In a reinforcement learning model, an agent interacts with the environment via perception and action, and a Markov Decision Process (MDP) is always used for reinforcement learning (as shown above). MDP consists of the tuple $(S, A, T, r,\gamma)$. We can thus organize the categorization following the Markov Decision Process, especially the elements of the MDP tuple. Specifically, we identify the attack and defense targets of state and action, environment and reward. The state $s$ and action $a$ refer to the state of the environment and the action of the agent in reinforcement learning, or an expression based on $s$ or/and $a$ (e.g., Q-value). The environment includes the transition matrix in MDP and surrounding environment situations. The reward aspect pertains to studies that aim at the reward function of MDP in reinforcement learning. The taxonomy is shown in figure \ref{table_fl}. 

\begin{table}
\sidewaystablefn%
\newcommand{\tabincell}[2]{\begin{tabular}{@{}#1@{}}#2\end{tabular}}
\tiny
\tabcolsep=0.06cm
\centering
\renewcommand\arraystretch{1.2}
\caption{Taxonomy. }
\label{table_fl}
\centering

\begin{tabular}{c|lll}
\hline
Section & Subsection & Target & Possible impact\\
\hline
\multirow{5}*{ \tabincell{c}{Security in \\reinforcement \\learning}} & \multirow{2}*{ \tabincell{l}{Security of state and action\\ in MDP}} & Action e.g.~\cite{13lee2020spatiotemporally} & \tabincell{l}{State, Action, Reward,\\ Environment, Policy} \\

& & State e.g.~\cite{25zhao2020blackbox}& \tabincell{l}{State, Action, Reward,\\ Environment, Policy}\\
\cline{2-4}

& \multirow{2}*{ \tabincell{l}{Security of environment\\ in MDP}} & Transition matrix e.g.~\cite{9chan2020adversarial}& \tabincell{l}{State, Action, Reward,\\ Environment, Policy}\\

& & Surrounding situations e.g.~\cite{27wang2020falsification}& \tabincell{l}{State, Action, Reward,\\ Environment, Policy}\\
\cline{2-4}

& \tabincell{l}{Security of reward function\\ in MDP} & Reward e.g.~\cite{8zhang2020adaptive}& \tabincell{l}{State, Action, Reward,\\ Environment, Policy} \\
\hline

\multirow{5}*{\tabincell{c}{ Privacy in \\reinforcement\\ learning}} & \multirow{2}*{ \tabincell{l}{Privacy of state and action\\ in MDP}} & Action e.g.~\cite{2vietri2020private}&  \tabincell{l}{State, Action, Reward,\\ Environment, Policy}\\ 

& & State e.g.~\cite{61cheng2022multi}&\tabincell{l}{State, Action, Reward,\\ Environment, Policy}\\
\cline{2-4}

&  \multirow{2}*{\tabincell{l}{ Privacy of environment\\ in MDP}} & Transition matrix e.g.~\cite{6pan2019you}& \tabincell{l}{State, Action, Reward,\\ Environment, Policy}\\

& & Surrounding situations e.g.~\cite{58ye2020differentially}& \tabincell{l}{State, Action, Reward,\\ Environment, Policy}\\
\cline{2-4}

& \tabincell{l}{Privacy of reward function\\ in MDP} & Reward e.g.~\cite{5liu2021deceptive}& \tabincell{l}{State, Action, Reward,\\ Environment, Policy}\\
\hline
\end{tabular}

\footnotetext{In this article, we first divide security and privacy in reinforcement learning into two main sections: security in reinforcement learning and privacy in reinforcement learning. We then further classify the related research from the perspective of MDP. Every main section will have three subsections: state and action, environment, and reward. These three aspects are all potential attack/defense targets, and may have some impact on the process. }
\end{table}

The agent's purpose is to find an optimal policy that can map environment states to agent actions. To facilitate more efficient policy evaluation, the concept of reward was introduced to reinforcement learning. A reward is a scalar signal that can be regarded as an indicator of the value for the state transition. Long-term rewards can be adopted to assess the policy; examples include the mean value of the reward, accumulated reward, or other functions based on rewards, such as the Q-value and V-value. 

At each step, the agent first observes the state $s_t$ from the environment. Next, the agent takes an action $a_t$ based on the policy $ \pi $; subsequently, the agent will receive a reward $r_t$ as feedback from the environment, and the environment changes to $s_{t+1}$. The transition matrix can be defined as a probability mapping from state-action pairs to states $T:(S\times A) \times S\rightarrow[0,1]$. It is usually difficult to determine an optimal policy based solely on the calculations of the evaluating function. Hence, value iteration methods are adopted to solve this issue. The action-state value iteration formula for evaluating the policy can be expressed as follows:
\begin{equation}
\begin{aligned}
    Q_{i+1} (s,a)=Q_{i}(s,a)+\alpha \left[r+\gamma \sum{\pi(a_{t+1} \vert s_{t+1})}Q_i(s_{t+1},a_{t+1})-Q_{i}(s_t,a_t) \right]
\end{aligned}
\end{equation}
where $\alpha$ is the learning rate; $s$ and $a$ are the state and action in current step $t$, and $s_{t+1}$ and $a_{t+1}$ are state and action in the next step.

\begin{figure}[htbp]
\centering
\includegraphics[width=0.9\textwidth]{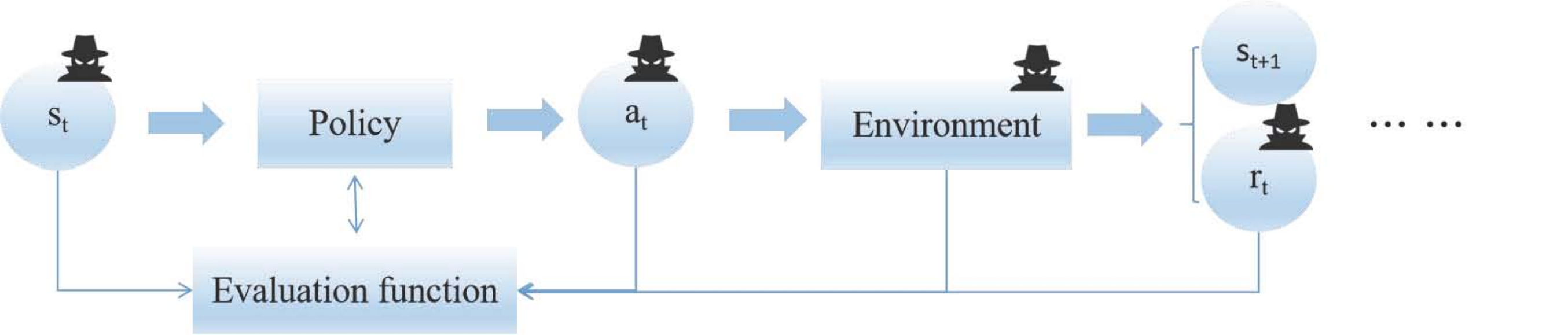}
\caption{The objects of attacks/defenses and their impacts on the reinforcement learning process. We can observe that all elements are situated in the chain of the reinforcement learning process; they are not isolated, but connected to each other. Therefore, attacks aimed at $s_t$, $a_t$, $r_t$, and the environment (such as the transition matrix $T$) can all affect the elements in the chain. }
\label{ai}
\end{figure}

Figure \ref{ai} illustrates the process, along with the attacks of the main elements classified in this paper. 

We can determine that the state $s_t$ of the environment may affect the action taken. In more detail, an agent using the policy output an action based on the state. The action $a_t$ will in turn affect the environment, and may impact the reward $r_t$ and the next state $s_{t+1}$. Moreover, the function used to evaluate the policy is based on $s_t$, $a_t$ and $r_t$. Consequently, if an adversary attacks $s_t$, the action $a_t$ may be directly influenced, while the reward and policy may also be impacted. If opponents attack $a_t$, the next state and reward may be changed by this attack; furthermore, the policy may also be affected. In addition, attacks targeting rewards can also influence the policy, while attacks on the environment (like the transition matrix, which is a probability mapping from state-action pairs to states) also have an influence on the choices of action and the next state. Thus, attacks aimed at $s_t$, $a_t$, $r_t$, and the environment (such as the transition matrix $T$) can all influence the agent to make a sub-optimal decision. In addition, the learning process is continuous; thus, attacks on every component in the chain may affect each other in some steps, and as a result, the agent may fail to achieve its goal.

\section{Security in reinforcement learning}
Security is one of the most significant aspects of reinforcement learning, as it involves exploring possible attacks and defenses to improve the robustness of the model. It tends to get a model which has a reliable and stable performance even when faced with sudden interference or changes. 

In this section, we will present a review of security in reinforcement learning based on the MDP perspective. This section can be divided into three subsections. The first addresses the security of the state and action in MDP. The second examines the security of the environment, which includes the transition matrix in MDP and the surrounding environment situations. Finally, we explore the security of the reward function in MDP. 

A summary of survey papers in this section is presented in Table \ref{table_zjs}. 
\begin{table*}[htbp]
\tiny
\tabcolsep=0.02cm
\newcommand{\tabincell}[2]{\begin{tabular}{@{}#1@{}}#2\end{tabular}}

\caption{Summary of research addressing security in reinforcement learning}
\label{table_zjs}
\centering
\renewcommand
\arraystretch{1.5}
\begin{tabular}{m{1.7cm}<{\centering}|l<{\centering}l<{\centering}l<{\centering}l<{\centering}l<{\centering}}
\hline
\rowcolor{lightgray}Subsection & Papers & Target& Impact & Strategies & \tabincell{c}{Representative\\
Methods}\\
\hline
\multirow{7}*{\tabincell{c}{Security of\\ state and\\ action \\ in MDP}} & Lee et al.~\cite{13lee2020spatiotemporally}  & Action	& Reward& Perturbations & \tabincell{l}{Optimization-based \\ approaches\\Projected gradient\\ descent}\\

& Chen et al.~\cite{14tessler2019action}  &	Action & Policy	& Action robustness	& \tabincell{l}{Zero-sum game\\Nash equilibriumt}\\

& Zhao et al.~\cite{25zhao2020blackbox}  & State & \tabincell{l}{Policy \\Action} & Perturbations & Imitation learning\\

& Garrett et al.~\cite{23garrett2019z} & State &  \tabincell{l}{System\\destabilization}  &	Perturbations &	Z tables\\

& Sun et al.~\cite{7sun2020stealthy}  & State & Reward, Action &	Perturbations &	\tabincell{l}{Prediction model\\Neural network}\\

&Ye et al.~\cite{99ye2022model} & State & Action & Model learning & Deep neural network\\

& Dai et al.~\cite{21dai2019reinforcement}  &	State-action & Policy &	Safe exploration & \tabincell{l}{Convolutional neural \\ network \\Transfer learning}\\
\hline

\multirow{7}*{\tabincell{c}{Security of \\ environment \\ in MDP}} & Rakhsha et al.~\cite{10rakhsha2020policy} &	 \tabincell{l}{Transition \\dynamics\\/ rewards } & Policy &	Data poisoning & \tabincell{l}{Optimization problems\\ having constraints}\\

& Chan et al.~\cite{9chan2020adversarial} &	Features & Reward & Adversarial sample	& \tabincell{l}{Sliding-window \\ method \\Gradient function}\\

& Wang et al.~\cite{27wang2020falsification} & \tabincell{l}{Environment\\ conditions} & Robust	policy  & \tabincell{l}{Robust adversarial\\ learning}  & \tabincell{l}{Cross-entropy\\ method \\Actor-critic \\ architecture}\\

& Li et al.~\cite{12li2019robust} & \tabincell{l}{	Non-stationary\\ environment} & Robust policies & \tabincell{l}{Robust adversarial\\ learning} & \tabincell{l}{Minimax optimization \\ End-to-end \\ learning approach}\\

& Lin et al.~\cite{26lin2020robustness} &	Features & Action &	Adversarial sample &	\tabincell{l}{Gradient-based \\ methods}\\

& Li et al.~\cite{103li2022policy}& Environment & Policy &  \tabincell{l}{Two-player \\zero-sum game} & Nash equilibrium\\

& Zhai et al.~\cite{104zhai2022robust}& Environment & Policy &  \tabincell{l}{Two-player \\zero-sum game} & \tabincell{l}{Nash equilibrium\\ Lyapunov network}\\

\hline

\multirow{2}*{\tabincell{c}{Security of\\ reward \\function\\in MDP}}& Zhang et al.~\cite{8zhang2020adaptive}  &	Reward &	Policy &	Poisoning attack&	\tabincell{l}{Optimal control\\ problems}\\
& Li et al.~\cite{11li2019dialogue} &	Reward &	Policy	 & \tabincell{l}{Adversarial inverse\\ reinforcement learning}
 & \tabincell{l}{Imitation learning\\ Entropy \\ regularization term}\\

\hline
\end{tabular}
\end{table*}

\subsection{Security of state and action in MDP }
The state and action in MDP are the state of the environment and the action of the agent in reinforcement learning. Through the use of adversarial examples or the application of sudden perturbations to these elements, a reinforcement learning model can be misled from optimal performance. Many existing methods focus on attacking the state and action in reinforcement learning; at the same time, methods of defending against these attacks have also been studied. 

In this section, we attempt to survey the security problems that target the state or/and action in MDP of reinforcement learning. Lee et al.~\cite{13lee2020spatiotemporally} and Chen et al.~\cite{14tessler2019action} focused on the action in MDP and studied the attacking and protecting of actions respectively. Zhao et al.~\cite{25zhao2020blackbox}, Garrett et al.~\cite{23garrett2019z}, Sun et al.~\cite{7sun2020stealthy} and Ye et al.~\cite{99ye2022model} all focused on attacking states in MDP of reinforcement learning. Moreover, Dai et al.~\cite{21dai2019reinforcement} learned to protect both the actions and states in MDP. 

\subsubsection{Action Security in MDP}
Attackers targeting actions can misdirect the agent's subsequent progress, and may also influence the reward or policy. One of the popular basic approaches to such attacks involves the use of adversarial samples that add perturbations to the action space. It is also possible to disturb the action sequence to produce inaccurate performance. 

Lee et al.~\cite{13lee2020spatiotemporally} studied strategies for attacking the action space in MDP of reinforcement learning. They proposed two novel attack strategies, both of which were optimization-based approaches. The first is a Myopic Action Space (MAS) attack, which creates perturbations in a greedy manner without future considerations and then distributes these attacks across the action space dimensions. The second is a Look-ahead Action Space (LAS) attack, in which the attacker can make predictions, selects a designed sequence of future perturbations, and distributes the attacks across the action and temporal dimensions. The results show that LAS attacks have a greater influence on the agent than MAS attacks. Projected gradient descent (PGD) is used for the two formed optimal problems. 

The approaches outlined above focus on attacking action space. Conversely, producing robust actions for reinforcement learning is also important and merits research attention.

Chen et al.~\cite{14tessler2019action} investigated reinforcement learning under action attack and considered actions to be robust in two situations. One is the Probabilistic Action Robust MDP (PR-MDP), with an adversary that adds a perturbation to the selected action; the other is the Noisy Action Robust MDP (NR-MDP), in which an agent has a probability of taking an alternative adversarial action. PR-MDP and the NR-MDP can both be viewed as a zero-sum game. Subsequently, Policy Iteration (PI) schemes are used to solve these problems in order to reach Nash Equilibrium or convergence. 

\subsubsection{State Security in MDP}
The state space, which is another important part of MDP, has a similar character to the action space in reinforcement learning systems. Both of them can affect the evaluation of policy and the environment. Methods of perturbing the state space or deceiving agents in this space have also been investigated, with particular attention paid to the dynamics of the agent cases. There are two common approaches to crafting state-space perturbations of reinforcement learning: model-based means and optimization based-means. 

Zhao et al.~\cite{25zhao2020blackbox} studied adversarial sample attacks in reinforcement learning using a model-based mean. They attempted to formulate an approximate sequence-to-sequence (seq2seq) model to predict a single agent action or a sequence of future agent actions, and produce adversarial samples to affect the state in MDP. They considered a full black-box attack situation, in which the attacker has no message of the agents, regardless of the training parameters or training methods of the agents. Finally, the adversarial samples are used to trigger a trained agent to misbehave at a specific time. This is a new concept in adversarial sample reinforcement learning that involves a time-bomb attack. 

For certain high-dimensional continuous state space situations, computing an optimization-based approach may be a more suitable means of generating attacks for a target agent compared to training another model. 

Garrett et al.~\cite{23garrett2019z} proposed an optimization method to formulate an attack that can directly influence the state(s). These attacks can affect the learning policy, cause the agent to act sub-optimally, and destabilize the cyber-physical system; this is made possible by the assumption that, in some cases, a foe can manipulate sensing/actuation signals. These authors used a Z table to conduct an attack. The Z table is a measure of how effective an adversarial effect to a state is versus the cost to perform the effect. Then using iteration to get the target state(s). 

Models also can be used as an auxiliary means of making predictions that improve the attack methods. Sun et al.~\cite{7sun2020stealthy} proposed adversarial attacks that add perturbation into the agent’s observation state. The goal of adversarial attacks against Deep Reinforcement Learning (DRL) is to inject a small set of adversarial samples in critical moments. For simplicity, this problem can be regarded as two sub-problems: that of "when to attack”, and that of “how to attack”. Critical point attack and antagonist attack are two models proposed to predict environment states and accordingly discover the critical steps and locations for perturbations. Existing adversarial example techniques are then used to compute and add perturbations in the selected critical moments. 

Ye et al.~\cite{99ye2022model} proposed a model-based self-advising method for multi-agent learning. This method enables the agents with the same ability when asked for advice on an unfamiliar state. The idea is to train a model based on states similar to a certain state. These authors provided a defined conception of "Similar States" and adopted a deep neural network to train a teacher agent on states both unfamiliar and familiar to the student. This method produces an improvement in learning and more robust performance with a much lower communication overhead.

\subsubsection{State and Action Security in MDP}
Sometimes, action and state space can be considered at the same time, at which point they are regarded as state-action pairs. These pairs can affect the reward function and potentially skew the policy so that the agent makes poor decisions in the Markov process. 

Dai et al.~\cite{21dai2019reinforcement} proposed a reinforcement learning method that considers safe exploration by evaluating the risk level of each state-action pair, then recording the most dangerous state-action pairs based on the security performance metrics in a blacklist. This algorithm employs a modified Boltzmann distribution based on the Q-values and the risk levels to choose an action. Moreover, a convolutional neural network was adopted to weigh the long-term risk levels of each state-action pair, while transfer learning was chosen to reduce the initial explorations in initial parameters learning.

\subsubsection{Summary and Discussion}
In this section, we surveyed the security problems focused on state and action in reinforcement learning.

We can observe that almost all the papers mentioned above consider adding perturbations to the target to influence a system. Moreover, the popular method of identifying a good perturbation is to regard the issue as an optimization problem. As a result, optimization-based approaches such as the gradient method and some intelligent approaches such as imitation learning have been selected for these purposes. Optimization-based approaches are traditional algorithms used to solve optimization problems, which are very efficient and intuitive; however, these methods may be useless for complex problems. In contrast, intelligent approaches can deal with these problems, thus these approaches can extend to more areas. Nevertheless, intelligent approaches may require more data for learning and more computing resources. 

\begin{table*}[htbp]
\sidewaystablefn%
\newcommand{\tabincell}[2]{\begin{tabular}{@{}#1@{}}#2\end{tabular}}
\tiny
\tabcolsep=0.04cm
\caption{Comparison of adding perturbations for state and action security}
\label{table_csas1}
\centering
\renewcommand\arraystretch{1.3}
\begin{tabular}{|c|c|c|c|c|c|c|c|}
\hline
 Paper &Target & Impact & Key Idea & \tabincell{c}{Representative\\ Methods}& Pros& Cons\\
\hline
 Lee et al.~\cite{13lee2020spatiotemporally}  & Action	& Reward & \multirow{2}*{\tabincell{c}{Algebra-based\\ optimization\\ approaches}}& \tabincell{c}{Optimization-based\\ approaches\\Projected gradient\\ descent} &\multirow{2}*{\tabincell{c}{More\\ accuracy}}& \multirow{2}*{\tabincell{c}{May be useless\\ in non-convex\\ problems}}  \\
\cline{1-3}
\cline{5-5}
 Garrett et al.~\cite{23garrett2019z} 	& State	 & \tabincell{c}{System\\destabilization}   & &Z Tables& &\\
\hline
 Zhao et al.~\cite{25zhao2020blackbox} &  State & \tabincell{c}{Policy \\Action}  & \multirow{2}*{\tabincell{c}{Intelligent\\ approaches}}& Imitation learning&\multirow{2}*{\tabincell{c}{More\\ intelligent}}& \multirow{2}*{\tabincell{c}{ May require \\more computing\\ resources}}    \\
\cline{1-3}
\cline{5-5}
 Sun et al.~\cite{7sun2020stealthy} & State & Reward, Action  & &	\tabincell{c}{Prediction model\\ Neural network}&   & \\
\hline
\end{tabular}
\end{table*}
Moreover, most of the methods require access to the states and/or actions, so that, the attacker can influence the reinforcement learning agent. However, it is impractical and limited attack capability in the real world. Considering the black box situation is the challenge of these attacks. Furthermore, states and actions are in the MDP chain, and considering this problem in a connected way is also a challenge. Studying the relationship between all the components in MDP will increase the targeted property of the security issues.

\subsection{Security of Environment}
In reinforcement learning systems, an agent interacts with the environment and then adjusts its learning results based on the information obtained from the environment. The environment includes the transition matrix in MDP and surrounding environment situations; examples of the latter include the surrounding architecture or road conditions encountered by the agent in the real world. Affecting the transition matrix in MDP and the surrounding environment situations can also influence the reward or state of environment input received by the agent; thus, it can mislead the agent into taking incorrect actions, and can also affect the policy.

Rakhsha et al.~\cite{10rakhsha2020policy} and Chan et al.~\cite{9chan2020adversarial} focus on studying attacks on the transition matrix and the features obtained from the environment respectively. Wang et al.~\cite{27wang2020falsification}, Li et al.~\cite{103li2022policy} and Zhai et al.~\cite{104zhai2022robust} focused on policy robustness in disturbed environmental conditions. All three of the above ideas are for single-agent areas. However, Li et al.~\cite{12li2019robust} and Lin et al.~\cite{26lin2020robustness} focused on the environment issue in multi-agent fields. Li et al.~\cite{12li2019robust} aimed at building robust policies under a non-stationary environment, while Lin et al.~\cite{26lin2020robustness} proposed attacks on the feature from the environment that mislead the agent into taking a target action.

\subsubsection{Environment security of a single agent}
The issue of an attacker who harms or poisons the learning environment was researched by Rakhsha et al.~\cite{10rakhsha2020policy}. These authors focused on data poisoning attacks that can manipulate the rewards or the transition matrix in the reinforcement learning (RL) algorithm, based on the understanding that reinforcement learning agents aim to maximize their average reward in undiscounted infinite-horizon settings. To fabricate the attacks, optimal methods with constraints and bounds were used. This paper considered online learning settings with poisoned feedback, in which the agent uses a regret-minimization mechanism to learn a policy while considering the different attack costs for online learning settings. In addition, it also studied offline settings in which the agent is formulating plans in a poisoned environment. 

In many cases, the surrounding environment information is complicated or high-dimensional, meaning that pre-management is required. Thus, the opponent can not only focus on the surrounding environment data directly, but can also aim at the data after some calculating steps (for example, extracting features). 

Chan et al.~\cite{9chan2020adversarial} studied the adversarial attack strategy against DRL by crafting an adversarial sample that perturbs the features to efficiently affect the cumulative reward. A static reward impact map is presented to measure the influence on the cumulative reward made by inputting features, that have slight changes. Then, using the "reward impact map" measures the importance of a feature; subsequently, select suitable actions. Finally, a crafted adversarial sample based on the gradient function attacks the sample by perturbing the features. 

The studies above focus on attacks. However, it is also important to develop a more reliable and stable reinforcement learning system for the surrounding situation. Accordingly, some approaches were developed and tested in a perturbed or poisoned environment. 

Wang et al.~\cite{27wang2020falsification} developed a mechanism that considers reinforcement learning with safety falsification methods. This framework is a falsification-based robust adversarial reinforcement learning (FRARL) framework that trains the policy in the new adversarial environment; thus, the system can perform as an adversarial reinforcement learning mechanism to enhance the robustness of trained policies. It is the first generic mechanism that combines temporal logic falsification with adversarial learning to improve policy robustness. A cross-entropy method is used to get the initial conditions and input sequences. 

Li et al.~\cite{103li2022policy} considered policy learning for Robust Markov Decision Processes (RMDP) from another unique perspective. Specifically, rather than concentrating on attack problems, these authors focused on the robustness of the environment to simulator domain mismatch in real application scenarios. They treated the mismatch as a perturbation and established the goal of finding a robust policy that ensures a near-optimal reward against the worst-case perturbation. A two-player zero-sum game was developed that considers the perturbation as an adversarial player, and Nash Equilibrium (NE) was used to find the robust policy. 

Zhai et al.~\cite{104zhai2022robust} also studied the problem of the differences between simulated and real environment, which may reduce the performance of the learned policies. These authors also modeled environmental differences as adversarial disturbances and constructed a two-player-zero-sum game between the normal and adversarial agents. However, such a method may increase the difficulty of the training domain. Consequently, this paper also considered certain constraints in the adversarial architecture and used a data-driven Lyapunov network to ensure the stability of the system during training.  

We can observe that single-agent reinforcement learning is delicate and sensitive in the training surrounding situation. Notably, this issue is even more pronounced in multi-agent scenarios. Researchers have accordingly studied these problems to make the learned models more robust for multi-agents. 

\subsubsection{Environment security of multi-agents}
Li et al.~\cite{12li2019robust} studied the problem of training robust deep reinforcement learning agents with continuous actions in the multi-agent learning setting. They proposed a Minimax Multi-Agent Deep Deterministic Policy Gradient algorithm (M3DDPG) to improve the robustness of the multi-agent reinforcement learning system. Their approach introduced the Minimax Optimization idea to update policies considering the worst situation. When this approach is applied, each agent operates under the assumption that all other agents are acting adversarially. The algorithm was based on Multi-Agent Deep Deterministic Policy Gradient (MADDPG), a decentralized policy and a centralized critic framework, used to obtain a Q function. These authors further proposed Multi-Agent Adversarial Learning (MAAL), which can approximate the non-linear Q function using a locally linear function to solve this optimization problem. 

Lin et al.~\cite{26lin2020robustness} studied attacks on cooperative multi-agent reinforcement learning (c-MARL), which is a necessary element of improving the robustness of the algorithm. In this study, perturbing agents' observations with an adversarial example can mislead the agents and minimize the value of a team's total reward. This special attack method comprises two steps. First, the adversary selects actions that can minimize the total team reward using a policy network trained with reinforcement learning. Next, the rival perturbs the agent’s observation by using targeted adversarial examples and gradient-based methods to make the agent take specific actions. 

\subsubsection{Summary and Discussion}
In this section, we surveyed the security problems focused on the environment in reinforcement learning. The environment is complex in the real world, and it is hard for attackers to lead a targeted attack.

Some of the papers discussed above focus on attacks, while others are about defending the environment. Attacks of this kind almost always utilize adversarial samples, while defenses against such attacks are always robust adversarial learning methods that consider training the agent in a worse situation to obtain a robust model. 

\begin{table*}[!htbp]
\newcommand{\tabincell}[2]{\begin{tabular}{@{}#1@{}}#2\end{tabular}}
\tiny
\tabcolsep=0.1cm
\caption{ Comparison attacks and defences for environment security }
\label{table_cse}
\centering
\renewcommand\arraystretch{1.3}
\begin{tabular}{|c|c|c|c|c|}
\hline
Strategies& Paper	 &Categories & Impact  &	Representative Methods\\
\hline
 \multirow{2}*{Adversarial sample} &Chan et al.~\cite{9chan2020adversarial}	 &	 Attack  & Cumulative reward 	& \tabincell{c}{Sliding-window method\\Gradient function}\\
\cline{2-5}
& Lin et al.~\cite{26lin2020robustness} &	 Attack & Action  &	Gradient-based methods\\
\hline
\multirow{2}*{ \tabincell{c}{Robust adversarial\\ learning}} & Wang et al.~\cite{27wang2020falsification}  & Defense  & Robust	policy   & \tabincell{c}{Cross-entropy method\\Actor-critic architecture}\\
\cline{2-5}
& Li et al.~\cite{12li2019robust} &  	Defense  & Robust policies  & \tabincell{c}{Minimax optimization \\End-to-end learning approach}\\
\hline
\end{tabular}
\end{table*}

As for the adversarial training methods, it is passive against attacks; because it injects adversarial examples into the training set to enhance the robustness of the model. So, it is useful only for considered adversarial examples.  Moreover, learning defense against adversarial attacks in black-box cases is also a challenge. Furthermore, in the security of the environment situation, if we want to have a targeted attack, the relationship of both all the components in MDP and other agents in multi-agent cases is a challenge.

\subsection{Security of reward function in MDP}
In many reinforcement learning (RL) applications, the agent extracts reward signals from the user or environment. A reward is crucial for a Markov decision process, as it indicates the feedback received by agents when they take certain actions in certain states. Ultimately, the reward is used to decide which action is optimal. Thus, adversaries tend to attack the reward functions of MDP in reinforcement learning. 

\subsubsection{Attack reward function in MDP}
The reward-poisoning attacks issued against reinforcement-learning agents were investigated by Zhang et al.~\cite{8zhang2020adaptive}. In this paper, these authors studied the training-time reward poisoning attack problem, which involves crafting environmental rewards and forces the reinforcement learning agent to learn a nefarious policy. They regarded the reward shaping task as an optimal control problem on a higher-level attack MDP. They also characterized conditions in situations where such attacks are useless, as well as provided upper bounds on the attack cost when an attack is feasible. 

The concept discussed above involves learning to attack the reinforcement learning systems through reward. A robust reward signal therefore becomes important, and has accordingly attracted significant attention from researchers. 

\subsubsection{Protecting reward function in MDP}
Li et al.~\cite{11li2019dialogue} focused on the problems of reward signal sparsity and instability in the field of dialogue generation. Dialogue reward learning with adversarial inverse reinforcement learning (DG-AIRL) is proposed to address this issue. This mechanism is a sequence-to-sequence (Seq2Seq) model that adopts adversarial imitation learning to enable the model to give human-like dialogue responses, and further designs a specific reward function structure to measure the reward of each word in the generated sentences. An entropy regularization term is also used to improve training stability. 

\subsubsection{Summary and Discussion}
The above-mentioned paper aimed at attacking the reward function using a poisoning attack that selected an optimal control method. It is possible that learning methods could be used to handle such a problem. However, these poisoning attacks also require access to the training data set, which is a limitation of the attack; thus, learning more practical methods is a challenge. Moreover, the adversarial sample which is another popular attack algorithm can also be applied to reward function attacks in future work. In addition, the adversarial inverse reinforcement learning method can successfully improve the quality of intelligence in reward inferring and may be extended in other areas.

\section{Privacy in reinforcement learning}

In this section, we will discuss privacy in reinforcement learning problems such as obtaining or inferring private information and protecting user privacy, with a particular focus on the MDP perspective. This section can be divided into three subsections. The first explores the privacy of the state and action in MDP. The second one addresses the privacy of the environment, which encompasses the transition matrix in MDP and the surrounding environment situations. The last part examines the privacy of the reward function in MDP. 

Table \ref{table_zjp} provides a concise summary of the survey papers in this section. 

\begin{table*}[!htbp]
\newcommand{\tabincell}[2]{\begin{tabular}{@{}#1@{}}#2\end{tabular}}
\tiny
\tabcolsep=0.02cm
\caption{Summary of research addressing the Privacy in reinforcement learning}
\label{table_zjp}
\centering
\renewcommand\arraystretch{1.3}
\begin{tabular}{l|lllll}
\hline
\rowcolor{lightgray}Subsection & Papers  &	Target & Impact &	Strategies &	Representative Methods\\
\hline
\multirow{6}*{\tabincell{l}{Privacy of \\ state and action \\ in MDP}} & Sakuma et al.~\cite{4sakuma2008privacy}  &	Q-values & \tabincell{l}{States, actions,\\ reward} & \tabincell{l}{	Cryptographic\\ solutions} &	\tabincell{l}{Additive homomorphic\\cryptosystem}\\

& Wang et al.~\cite{3wang2019privacy}	& Q-values & Reward &	Differential privacy	& \tabincell{l}{Gaussian process noise\\Common neural networks}\\

& Ye et al~\cite{60ye2020differential} &	Q-values &	Policy&	\tabincell{l}{Differential \\ advising method} &	Laplace mechanism\\

& Cheng et al.~\cite{61cheng2022multi} &	Q-values &	Policy &	\tabincell{l}{Differential transfer\\learning method} &	\tabincell{l}{Differentially \\ exponential noise\\Relevance weight}\\ 

& Vietri et al.~\cite{2vietri2020private}  & States, actions & \tabincell{l}{States, actions\\reward} & Differential privacy & \tabincell{l}{Joint differential privacy\\ Optimistic strategy} \\

&\tabincell{l}{Chowdhury and\\ Zhou~\cite{102chowdhury2021differentially}} & States, actions & \tabincell{l}{Transition\\ probabilities} & Differential privacy & \tabincell{l}{Joint differential privacy\\ Local differential privacy}\\

\hline
\multirow{3}*{\tabincell{l}{Privacy of \\ environment \\ in MDP}} & Pan et al.~\cite{6pan2019you}  &\tabincell{l}{ Transition\\ dynamics matrix} &	\tabincell{l}{Surrounding\\ situations}  &	\tabincell{l}{Environment \\ dynamics search}  & \tabincell{l}{Genetic algorithm \\Shadow policies}\\

& Ye et al.~\cite{58ye2020differentially}  &	Environment	& Q-values,reward &	Differential privacy & \tabincell{l}{Laplace noise \\Privacy budget}\\	

& Zhou et al.~\cite{68zhou2022differentially}  & Environment & Policy & \tabincell{l}{Joint differential\\ privacy} & Gaussian mechanism\\ 
\hline

\multirow{3}*{\tabincell{l}{Privacy of \\ reward function \\ in MDP}} & Liu et al.~\cite{5liu2021deceptive}  &	Reward function & Q-values &	 \tabincell{l}{Reinforcement\\ learning} &	\tabincell{l}{Ambiguity model\\ Intention recognition}\\

& Ye et al.~\cite{59ye2019differentially}  &	Reward function &	Policy &	Differential privacy &	Laplace noise\\

& Fu et al.~\cite{69fu2017learning}  & Reward & Policy & Adversarial learning & \tabincell{l}{Inverse reinforcement\\  learning}\\

\hline
\end{tabular}

\end{table*}

\subsection{Privacy of state and action in MDP}
The agent and the environment interact through states and actions, which may be stolen by adversaries and cause privacy leakage problems.  

States and actions in MDP encompass the state of the environment, the action of the agent, and other concepts based on state and action. Vietri et al.~\cite{2vietri2020private} aim at protecting the state and action directly in MDP. Wang et al.~\cite{3wang2019privacy}, Sakuma et al.~\cite{4sakuma2008privacy}, Ye et al.~\cite{60ye2020differential}, Cheng et al.~\cite{61cheng2022multi}, and Chowdhury and Zhou~\cite{102chowdhury2021differentially} focused on protecting the Q-functions based on states and actions. 

\subsubsection{Using differential privacy to ensure the privacy of state and action in MDP}
For privacy protection problems, differential privacy is a popular model that has been widely used in many areas. Differential privacy considers the worst-case situation in which attackers know all the data except for a new record in a dataset. It can ensure that any individual record being stored in or removed from a dataset makes little difference to the dataset's analytical output~\cite{28arulkumaran2017deep}. 

The work of Wang et al.~\cite{3wang2019privacy} considered the use of differential private methods for reinforcement learning in continuous spaces, with a focus on protecting the value function approximator. The algorithm added Gaussian process noise to the corresponding action-state value function of deep Q-learning, which can satisfy differential privacy guarantees at every iteration. It also chose the reproducing kernel Hilbert space (RKHS) embedding common neural networks for the nonlinear value function. 

Ye et al.~\cite{60ye2020differential} also adopted differential privacy to study issues in reinforcement learning. These authors proposed a novel differential advising in multi-agent reinforcement learning inspired by the differential privacy mechanism. Using this approach, an agent can take advice produced with reference to a slightly different state. It also added Laplace noise to the agent advice Q(s), which can provide the agent with more reliable data for use in decision-making. 

The knowledge transfer problem in multi-agent reinforcement learning was studied by Cheng et al.~\cite{61cheng2022multi}. They proposed a Differential knowledge Transfer with relevance Weight (DTW) algorithm, which also chose differential privacy to handle this problem. DTW was embedded into the multi‐agent reinforcement learning algorithm to add differential noise and relevance weights to the Q‐value. This model can expand the knowledge set and reduce the influence of negative transfer. 

Far less attention has been paid to addressing privacy in reinforcement learning problems, compared with private bandit algorithms. Vietri et al.~\cite{2vietri2020private} proposed the first reinforcement learning algorithm for regret minimization with the JDP guarantee. These authors designed an algorithm named the Private Upper Confidence Bound algorithm (PUCB) for reinforcement learning, which used differentially private guaranteeing of the Laplace mechanism to protect the information of training data, and moreover imposed lower bounds on the regret and a smaller number of sub-optimal episodes. It computed the policy of the reinforcement learning algorithm using private counts such as $\hat{n} (s,a,h)$ and $\hat{m} (s,a,s^{'},h)$, which are the number of times the agent has taken action $a$ in state $s$ at time $h$ and the number of times the agent has taken action $a$ in state $s$ at time $h$ and transitioned to $s^{'}$ respectively. This method satisfies the joint differential privacy (JDP) with lower-bounds sample complexity and regret of probably approximately correct (PAC). 

Chowdhury and Zhou~\cite{102chowdhury2021differentially} also considered the regret bounds of reinforcement learning. They proposed two general frameworks for designing private, optimistic reinforcement learning algorithms, one for policy optimization and another for value iteration, that also satisfied JDP and LDP requirements. They designed the counts returned by the privatizer, which depend on users’ states and actions to calculate the private mean empirical costs and private empirical transitions. Examples include the count $N^{k}_{h}(s,a)$ and its privatized versions $\tilde{N}^{k}_{h}(s,a)$, which denote the number of times that the agent has visited state-action pair $(s,a)$ at step h before episode k and are similar to the counts in~\cite{2vietri2020private}. The frameworks also can obtain sublinear regret guarantees.

\subsubsection{Using cryptography to improve the privacy of state and action in MDP} 
Sakuma et al.~\cite{4sakuma2008privacy} studied the issue of privacy in distributed reinforcement learning (DRL), devising privacy-preserving reinforcement learning algorithms using an additive homomorphic cryptosystem. The Q-values are encrypted, allowing the addition of encrypted values without requiring their decryption. Data partitioning by time and by observation were considered, and random action selection was used for these two aspects.  the classical privacy-preserving method cryptography is still popular. 

\subsubsection{Summary and Discussion}
We surveyed privacy of state and action in MDP in this section and found that even in real-world applications, there are many methods like some tools or frameworks such as Privacy-Preserving and Security Mining Framework (PPSF)~\cite{109lin2018ppsf} in privacy-preserving problems, the most popular methods in this area are Differential privacy and cryptography.

Differential privacy and cryptography are both prevalent privacy models, each with its advantages and disadvantages. Differential privacy may be more flexible and can be used in many fields due to its ability to achieve a balance between data utility and privacy. However, this method may sometimes be unreliable, leading to privacy leakage and the sacrifice of some data utility. In contrast, cryptographic techniques are very reliable, but at the same time, they require a great deal of calculation and may reduce the efficiency of data sharing. The challenge in this situation is to choose a method which can balance the data utility and privacy. 

\begin{table*}[!htbp]
\newcommand{\tabincell}[2]{\begin{tabular}{@{}#1@{}}#2\end{tabular}}
\tiny
\tabcolsep=0.04cm
\caption{Comparison of methods for defending the privacy of state and action in MDP}
\label{table_cpas}
\centering
\renewcommand\arraystretch{1.3}
\begin{tabular}{|l|l|l|l|l|l|}
\hline
Key idea & Papers&	Target   &	Representative Methods& Pros & Cons\\
\hline
\tabincell{c}{Cryptographic \\solutions} & Sakuma et al.~\cite{4sakuma2008privacy}  &		Q-values   &	\tabincell{c}{Additive homomorphic\\cryptosystem}& Reliable & \tabincell{c}{High resource \\consumption\\ Low data sharing\\ efficiency} \\
\hline
\multirow{5}*{\tabincell{c}{Differential \\privacy}} & Wang et al.~\cite{3wang2019privacy} &  Q-values  	& \tabincell{c}{Gaussian process noise\\Common neural networks}& \multirow{4}*{Efficient}&\multirow{4}*{\tabincell{c}{Unsafe in\\ some cases}}\\
\cline{2-4}
& Ye et al.~\cite{60ye2020differential}	 &	Q-values  &	Laplace mechanism& & \\
\cline{2-4}
& Cheng et al.~\cite{61cheng2022multi} &	Q-values &	\tabincell{c}{Differentially exponential noise \\Relevance weights}& & \\ 
\cline{2-4}
& Vietri et al.~\cite{2vietri2020private} &  \tabincell{c}{States\\ actions}  & \tabincell{c}{Joint differential privacy\\ Optimistic strategy}& &  \\
\cline{2-4}
& \tabincell{c}{Chowdhury and\\ Zhou~\cite{102chowdhury2021differentially}} & \tabincell{c}{States\\ actions} & \tabincell{c}{Joint differential privacy\\ Local differential privacy}& & \\

\hline
\end{tabular}
\end{table*}

\subsection{Privacy of environment}
Interaction and data sharing occur frequently between the agent and the environment, and the environment is closely connected with other information in MDP. Hence, there is also a high probability of environmental privacy leakage. Opponents will focus on stealing the transition matrix data of the environment or information of the surrounding conditions (The surrounding conditions denote the area in which the agents are trained and/or to which they are applied). 

\subsubsection{Researches in privacy of environment}
The privacy leakage problem in deep reinforcement learning was studied by Pan et al.~\cite{6pan2019you}. These authors focused on the problem of environment dynamics search with the goal of inferring the environment. Genetic algorithm and shadow policies were selected for optimal policy selection and candidate inference respectively. This private environment information leaking problem was considered under two different scenarios. In the first, the attacker has no knowledge about the training surrounding situations, and the environments just with common constraints. In the second, the attacker has access to a set of potential candidates of the training environment dynamics. 

Ye et al.~\cite{58ye2020differentially} studied private information leakage of Multi-Agent planning for logistic-like problems. These authors adopted a planning approach that employed a reinforcement learning algorithm to make a plan to find the optimal route from the initial state to the goal state. This paper, proposed an approach adopting the differential privacy technique to achieve strong privacy preservation in multi-agent environments. It also used the concept of a “privacy budget ” to control the communication overhead. 

Zhou et al.~\cite{68zhou2022differentially} proposed to protect users’ sensitive and private information by considering regret minimization in large state and action spaces. Their work used the notion of joint differential privacy (JDP) and considered MDPs by means of linear function approximation. It further proposed two algorithms that applied the Gaussian mechanism to the information of the environment to protect the features. The two proposed privacy reinforcement learning algorithms are based on value iteration and policy optimization. 

\subsubsection{Summary and Discussion}
We surveyed the papers about the privacy of the environment in recent years in this section. Maybe because of the complexity of the environment, there are not been so many studies in this area recently. For the complexity of the environment, learning the relationship between the environment and other components in MDP is a challenge, and it can help us to comprehend the process of reinforcement learning and to generate a targeted and efficient attack. Moreover, the popular method of defense against attacks on privacy is still differential privacy and its variation. The attack method on the privacy of the environment is an intelligent method, for the complexity of the environment, training the attack model considering the connections of all the components in MDP is a challenge. We may adopt GNN to train in the future. 

\subsection{Privacy of reward function in MDP}
Reinforcement learning is a framework within which an agent learns a behaviour policy through interacting with the environment and responding to positive and negative rewards~\cite{44sutton1998reinforcement}. The reward function always determines the amount of the reward and when it is given. Thus, the reward function is a key element of giving the reward, and is accordingly very important for reinforcement learning. However, it is likely that observers can attempt to infer information about the policy from the reward function. 

Liu et al.~\cite{5liu2021deceptive} aimed to preserve the privacy of a reward function in reinforcement learning. Specifically, these authors proposed two methods with more general dissimulation models to preserve reward privacy: the Ambiguity Model, in which the agent selects actions that maximize the entropy based on ambiguity, and the intention recognition model, which takes action selection as a weighted sum of honest and ‘irrational’ behaviour. These methods both use pre-trained Q-functions and produce a policy that makes it hard to use inverse reinforcement learning or imitation learning. Thus, an observer is difficult to obtain the reward function. 

Ye et al.~\cite{59ye2019differentially} addressed issues with Multi-agent Advising Learning and applied reinforcement learning to deal with the packet routing problem. These authors adopted differential privacy to reduce the impact of the malicious agent by adding Laplace noise to the accumulated reward to protect the information of each agent. The privacy budget concept was then used to control the communication overhead, which can improve the learning performance. 

Fu et al.~\cite{69fu2017learning} proposed an adversarial inverse reinforcement learning algorithm to acquire a robust reward for changes in dynamics. Inverse reinforcement learning focuses on the problem of inferring an expert’s reward function from demonstrations, and this paper combines such a mechanism with adversarial learning, which can improve the robustness of the algorithm. This approach enables the proposed algorithm to learn policies even in environments that undergo great changes during training; thus, it achieves better performance than prior IRL methods in continuous, high-dimensional situations with unknown dynamics. 

As differential privacy is always adopted to establish a mathematical way of guaranteeing data privacy in reinforcement learning, and considering that inverse reinforcement learning is applied to inferring the reward function from demonstrations and providing rewards to the learning system, Prakash et al.~\cite{101prakash2021private} investigated the existing set of privacy techniques for reinforcement learning and proposed a new Privacy-Aware Inverse reinforcement Learning (PRIL) analysis framework, which is a new form of privacy attack that targets the private reward function. This reward attack attempts to reconstruct the original reward from a privacy-preserving policy (such as differential privacy) using an inverse reinforcement algorithm. The results showed that privacy in the policy domain does not translate to privacy in the reward domain, as the reconstruction error is independent of the $\epsilon$-DP budget.

\subsubsection{Summary and Discussion}
Privacy problems related to the reward function are discussed. We can observe that differential privacy is the most popular method, and moreover that some intelligent algorithms have been developed in the privacy area. Intelligent algorithms can handle more complex problems and situations, even if little information about the model and the goal is available; thus, they can expand the horizons of privacy protection. We can further observe that inverse reinforcement learning has been adopted to address these privacy problems~\cite{68zhou2022differentially,69fu2017learning}. In fact, however, inverse reinforcement learning is more frequently used to approach the problem of inferring an expert’s reward function from demonstrations and to provide the reward to the learning system rather than to tackle security and/or privacy issues. However, the concept of "inferring a reward function" naturally prompts thoughts of privacy leakage and unreliable models. Consequently, it might be possible to consider the security and privacy problems of inverse reinforcement learning, like the work in \cite{101prakash2021private}.

\begin{table*}[!htbp]
\newcommand{\tabincell}[2]{\begin{tabular}{@{}#1@{}}#2\end{tabular}}
\tiny
\tabcolsep=0.02cm
\caption{Comparison of methods for defending environment and reward function privacy in MDP}
\label{table_cpr}
\centering
\begin{tabular}{|c|c|c|c|c|c|c|}
\hline
Key idea & Papers&	Impact&	Strategies	& \tabincell{c}{Representative\\Methods} & Pros & Cons\\
\hline
\multirow{2}*{\tabincell{c}{Intelligent\\ means}}& Liu et al.~\cite{5liu2021deceptive}   & Q-values &	\tabincell{c}{Reinforcement\\ learning} &	\tabincell{c}{Ambiguity model\\Intention recognition} &\multirow{2}*{\tabincell{c}{More\\ intelligent}}& \multirow{2}*{\tabincell{c}{Higher computati-\\onal consumption;\\ can only defend\\ against certain\\ specific attacks}}\\
\cline{2-5}
& Fu et al.~\cite{69fu2017learning}   & Policy & \tabincell{c}{Adversarial\\ learning} & \tabincell{c}{Inverse\\ reinforcement\\ learning} & &\\
\hline
\multirow{3}*{\tabincell{c}{Differential\\ privacy}}& Ye et al.~\cite{58ye2020differentially} 	& \tabincell{c}{Q-values\\
Reward} &\tabincell{c}{	Differential\\ privacy} & \tabincell{c}{Laplace noise \\Privacy budget}  &\multirow{3}*{\tabincell{c}{Balance the\\ data utility\\ and privacy}}&\multirow{3}*{\tabincell{c}{Higher\\ communication\\ overhead}}\\	
\cline{2-5}
& Zhou et al.~\cite{68zhou2022differentially}   & Policy &\tabincell{c}{ Joint differential\\ privacy} &\tabincell{c}{ Gaussian\\ mechanism} & &\\
\cline{2-5}
& Ye et al.~\cite{59ye2019differentially}   &	Policy &\tabincell{c}{	Differential\\ privacy} &Laplace noise & &\\
\hline
\end{tabular}
\end{table*}

\section{Security and privacy in reinforcement learning applications}
Besides the security and problems in reinforcement learning itself, there are several works about applying reinforcement learning to security and privacy problems.

Researchers have applied reinforcement learning to many areas to help increase the model performance. Chen et al.~\cite{112chen2021rdrl} applied reinforcement learning to the reconfigurable wireless network. The authors proposed a primary-prioritized recurrent deep reinforcement learning algorithm for dynamic spectrum access. In this work, the spectrum Markov state is modelled to capture the evolution behavior to achieve the priority queuing of the primary users and the secondary users, and Dueling DQN is used for dynamic spectrum access allowing the secondary users to modify their parameters to select the optimal access policy. Chen et al.~\cite{113chen2022game,114chen2021edge,115chen2021deep} also used reinforcement learning mobile edge computing environment. They studied a polling callback energy-saving offloading strategy to the time-sharing mobile edge computing data transmission problem. This strategy adopts Dueling DQN as part of the approximator to improve the ability of processing and predicting time intervals and delays in time series~\cite{113chen2022game}. They also applied a deep reinforcement learning offloading model to acquire network resource allocation and optimally offloading decisions in Convergence of Augmented Reality (AR) and Next Generation Internet-of-Things (NG-IoT) areas ~\cite{114chen2021edge}. They also considered deep reinforcement learning to improve the fog resource provisioning performance of mobile devices in the mobile edge computing (MEC) paradigm~\cite{115chen2021deep}. The use of reinforcement learning in the above areas makes full use of reinforcement learning to improve the security of the systems. Moreover, there also are many applications of reinforcement learning to help preserve privacy.

Belhadi et al.~\cite{107belhadi2021privacy} adopt reinforcement learning to for faults detection in the smart grid. The authors develop a new framework to identify anomalous patterns in a distributed and heterogeneous energy environment by reinforcement learning method with blockchain. In this framework, the data mining model is used to
discover the local outlier factor that can be used to find the generic
anomalous patterns locally. Then, a reinforcement learning model with blockchain is used to merge the locally generic anomalous patterns forming the global complex anomalous patterns and ensuring the security of the collected time series at the same time. 

Ahmed et al.~\cite{108ahmed2021privacy} considered reinforcement learning of privacy-preserving in vehicle Adhoc networks. In this work, a deep reinforcement learning method is used to sensitize the private information for a given vehicle connected over Vehicle Adhoc networks. The deep learning method is adopted to check adversarial attack compatibility while training and testing different architectures, which can improve and learn patterns of the sensors. The authors further proposed a privacy-preserve method for data mining in 5G environments~\cite{110ahmed2022deep}. This method combines entropy-based learning with an attention-based approach. It can effectively hide sensitive patterns.

Ren et al.~\cite{111ren2022privacy} adopt reinforcement learning to the Internet of Things (IoT) and proposed a novel Privacy-protected Intelligent Crowdsourcing scheme based on Reinforcement Learning (PICRL). The proposed PICRL can guarantee the data quality by an effective trust evaluation mechanism (evaluates the trust of participants) and the reinforcement method Q-learning is utilized to select participants and maximize the utility of the system.

Liu et al.~\cite{116liu2022distributed} focused on the Mobile edge computing (MEC) problem and proposed a privacy-preserving distributed deep deterministic policy gradient (P2D3PG) algorithm. This algorithm converts the distributed optimization problem which is about maximising the cache hit rate of all the cache entities in the MEC-enabled system into a distributed model-free Markov decision process (MDP) problem. Then distributed reinforcement learning method is used to solve these distributed problems.
Gao et al.~\cite{117gao2022ppo2} also studied the edge computing problem. The authors also adopt a Markov decision process (MDP) to model the process of solving an optimal task offloading decision problem and then, the deep reinforcement learning (DRL) method is also used to solve the planning problem considering the location privacy requirement.

\section{Discussion and future works}

In this paper, we survey the security and privacy problems in reinforcement learning from the perspective of MDP. Developing a stable and reliable algorithm is an important direction of reinforcement learning. Attacks and defences of MDP in this area are both key elements.

\subsection{Challenges}
Most machine learning methods are data-driven. Therefore, based on the perspective of data, there are three main challenges of security and privacy in machine learning. These are the preservation of data privacy, increasing the model robustness, and facing the emergence of distributively processed data~\cite{121rodriguez2022survey}.
We can compare the challenges of reinforcement learning, deep learning, and federated learning to help comprehend the challenges of reinforcement learning. As illustrate in table~\ref{table_cg}.
\begin{table}
\sidewaystablefn%
\newcommand{\tabincell}[2]{\begin{tabular}{@{}#1@{}}#2\end{tabular}}
\tiny
\tabcolsep=0.09cm
\centering
\renewcommand\arraystretch{1.5}
\caption{Comparison of main common challenges of three machine learning methods. }
\label{table_cg}
\centering

\begin{tabular}{|c|c|c|c|}
\hline
Challenges & Reinforcement learning & Deep learning & Federated learning\\
\hline
Preservation of data privacy & Yes & Yes & Yes\\
\hline
Increase the model robustness & Yes & Yes & Yes \\
\hline
Facing the distributively process data & No & Yes & Yes\\
\hline
\end{tabular}
\end{table}

Reinforcement learning, deep learning, and federated learning are all faced with preservation of data privacy issues for all of them need data for training. However, there is some difference. The main applications of deep learning are image classification. Thus, most of the privacy of the data is about image. As for reinforcement learning and federated learning, the privacy data rely on the environment they applied. As for the increase in the model robustness, reinforcement learning needs to interact with the environment to get knowledge. It is a dynamic process, therefore, the attacks and defences have more challenges about dynamic than other methods. For the distributively processed data part, federated learning focuses more on a distributed approach to tackle local and global learning, which traditional deep learning is not concerned about. However, reinforcement learning in a multi-agent situation has similar problems in tackling local and global learning.

Moreover, reinforcement learning is a process. The mostly used MDP contains four elements (state, action, reward and environment), and all these elements are studied by researchers. The researchers have taken different methods and considered different situations to study security and privacy issues in recent years. However, only focusing on each element is not enough. Reinforcement learning is a process, and every component in the process chain can affect the process and even impact the overall results. Moreover, some methods may be inefficient ignoring the relationship of the elements. for example, a small perturbation in a state may make the policy choose an action $a^{'}$ instead of $a$. However, $a^{'}$ may have the same reward as $a$. So, the attack is inefficient. In addition, reinforcement learning needs to interact with the complex environment to gather knowledge, and the agent may be influenced by the environment easily. Hence, considering the correlations (the internal components and external factors) is a challenge.

\subsection{Common security and privacy models}
Furthermore, we can find that the popular attack methods in reinforcement learning are adversarial attack, poisoning attack, genetic algorithm and inverse reinforcement learning methods from the surveyed paper. Some of these methods are similar to the attack methods in other machine learning areas such as Deep Learning and Federated Learning. For example, the poisoning attack is used in all these learning fields. All these learning fields need a lot of data for training, hence data poisoning can naturally be used to attack their training process. However, as federated learning allows users to collaboratively compute a global machine learning model based on user-specific local models, without revealing the users’ local private data, In general, the model poisoning attack can be used in federated learning. The model poisoning attack has a malicious party which can modify the updated model before sending it to the central server. Membership inference attacks which aim to get information by checking if the data exists on a training set are always used in federated learning. Attackers make use of the global model to get information about the training data details of the other users. Membership inference attacks also can be used in deep learning. These two attack methods are hardly used in reinforcement learning. However, the speciality in cooperation may be applied to the cooperated multi-agent reinforcement learning system. Inverse reinforcement learning (IRL) is another kind of inferring algorithm that can be used to infer the reward function of reinforcement learning based on the policy or observed behaviour~\cite{89arora2021survey}. It is mostly used in reinforcement learning currently.

Deep learning focuses more on training compared with reinforcement learning which is always "online" learning interacting with the environment. Therefore, attacks aimed at models are used in deep learning security and privacy problems such as model extraction attacks. Model extraction attack attempts to duplicate a learning model without prior knowledge of training data and algorithms. Dynamic attacks are more suitable for reinforcement learning which has a non-fixed policy in most situations. In conclusion, some of the attacks on reinforcement learning are the same as the attacks in other machine learning areas, and some of the attacks work based on the characteristics of reinforcement learning (such as inverse reinforcement learning attacks). We summarize some attacks in table \ref{table_md}.

\begin{table}
\sidewaystablefn%
\newcommand{\tabincell}[2]{\begin{tabular}{@{}#1@{}}#2\end{tabular}}
\tiny
\tabcolsep=0.09cm
\centering
\renewcommand\arraystretch{1.5}
\caption{Comparison of generic models of three machine learning methods. }
\label{table_md}
\centering

\begin{tabular}{|c|c|c|c|}
\hline
Generic models & Reinforcement learning & Deep learning & Federated learning\\
\hline
Adversarial attack & Yes & Yes & Yes\\
\hline
Data poisoning attack & Yes & Yes & Yes \\
\hline
Model poisoning attack & No & No & Yes\\
\hline
Membership inference & No & Yes & Yes\\
\hline
Inverse reinforcement learning & Yes & No & No\\
\hline
\end{tabular}
\end{table}

\subsection{Complexity of the security and privacy methods}
As the comparison in tables~\ref{table_csas1}, \ref{table_cpas}, and \ref{table_cpr}, researchers take methods to protect privacy and defense against attacks, the methods will require different computational consumption. Reinforcement learning is a dynamic process, the storage issues are not so obvious. Researchers are many focused on computational consumption (or time complexity). Some surveyed papers considered with the computational complexity analyse while some papers are only concerned the effect of the methods. Wang et al.~\cite{3wang2019privacy}, considered differential private algorithms for reinforcement learning in continuous spaces, which can protect the reward information from being exploited. The proposed method  can respond to $N_{q}$ queries in $O(N_{q} ln(N_{q}))$ time. Park et al.~\cite{1park2020privacy}  consider using homomorphic encryption (HE) scheme to propose a privacy-preserving reinforcement learning (PPRL) framework for the cloud computing platform. In this work, computational complexity of the user is given by $O(N_{s} · N log q · (N + R_{e}) + N_{a} · N · R_{d} )$ and of the cloud platform is given by $O(N_{s} · N_{a} · N^{2} log q + N · (N_{s} log q · R_{d} + N_{a} · R_{e}))$. where $N_{s}$ and $N_{a}$ are the number of states and actions. $R_{e}$ and $R_{d}$ are the integer numbers that can be determined based on the public key and private key. $q = Lp$, where $L$ is a parameter for encryption and $p$ is the cardinality of the plaintext set. $N$ is the vector size of ciphertexts.

As for the attack, some work uses  $l_{p}$-norm of differences to represent the Cost of the attack. Such as rakhsha et al.~\cite{10rakhsha2020policy}, studied a security threat to reinforcement learning where an attacker can manipulate the rewards or the transition dynamics. The cost of the attack can be defined as:

\begin{equation}
    \Vert \hat{R}- \overline{R} \Vert_{p}=( \sum_{s,a} {(\lvert\hat{R}(s,a)- \overline{R}(s,a)\rvert)^{p} } )^{1/p}
\end{equation}

\begin{equation}
     \Vert \hat{P}- \overline{P} \Vert_{p}=( \sum_{s,a}{ \sum_{s^{'}} {(\lvert \hat{P}(s,a,s^{'})- \overline{P}(s,a,s^{'}) \rvert)^{p} } } )^{1/p}
\end{equation}

where $\overline{R}$ is the original reward function and $\overline{P}$ is the original transition matrix. $\hat{R} $ is a poisoned reward function and $\hat{P}$ is a poisoned transition matrix.

\subsection{Future work}
There are numerous related research avenues that could be pursued in future.

\textbf{Attacking various components of MDP} 
In the papers discussed above, opponents may attack only one or two elements of MDP. We may study attacking many different elements in MDP in future, along with the impact caused by more elements. We could also investigate the combined action of attacking multiple elements simultaneously; it is possible that the combined effect will exceed the total impact of separate attacks. 

In addition, we can further study the relationship between the elements with the hope of more strongly influencing the system performance. We can infer other information about the system elements based on some elements in MDP, or find the most efficient attack point or moment based on the relationship between these elements. 

Furthermore, we could also discuss issues related to the limited knowledge possessed by attackers. For example, if an attacker possesses partial state information and partial Dynamic Transition probability information, it may be possible to produce attacks based on this limited knowledge alone and then combine the analysis of the relationship between these two elements. 

\textbf{Improving robustness of reinforcement learning} 
In real situations, the environment is complex and continuously changing. We can consider fault-tolerant control to develop a more robust algorithm that is capable of handling this complex and changing environment. For example, we could consider the situation in which the agent receives incorrect information, or fails to receive some part of the required information, because of issues with the sensors. We could also consider the situation in which the connection quality is poor, preventing the agent from receiving information in a timely fashion and consequently causing it to make sub-optimal decisions. 

An adaptive fault-tolerant control (FTC) approach for MIMO nonlinear discrete-time systems was proposed by Liu et al.~\cite{78liu2016adaptive}. In this paper, abrupt faults and incipient faults are both taken into account. Li et al.~\cite{79li2020adaptive} also focused on the adaptive fault-tolerant tracking control problem, and further considered the influence of the dead zones and actuator faults on the control performance. Future work could thus take specific faults into account based on the actual situation. 

\textbf{Security and privacy issues of smart methods in reinforcement learning}
Many intelligent methods have been used for reinforcement learning problems. These methods may be used in complex problems in which it is difficult to obtain the goal or the model. 

Inverse reinforcement learning (IRL) is one such intelligent method that is used to infer the reward function of the reinforcement learning model. Its application fields include video games, in which it is often more difficult to design a reward function that describes the behaviors and yields an optimal policy than to provide demonstrations of the target behaviour in the video games. Tucker et al.~\cite{75tucker2018inverse} used inverse reinforcement learning algorithms to infer a reward from demonstrations; this approach utilized CNNs to deal with high-dimensional video games. Neu et al.~\cite{76neu2009training} investigated the application of IRL algorithms to parser training problems, and were able to automatically find a reward function that matched the training set. IRL may be a good method for use in building a better reinforcement learning model, as it enables rewards to be obtained automatically. However, it also may be used to attack a system to infer the reward function. Moreover, to obtain rewards, IRL also needs data to train; as a result, privacy leakage problems may arise~\cite{101prakash2021private}. 

In addition, while such intelligent algorithms can improve the handling of certain problems, it might be preferable to focus on specific convergence analysis rather than simply learning in future work (for example, studying the regret bounds of reinforcement learning~\cite{102chowdhury2021differentially}).

\section{Conclusion}

In this paper, we investigated security and privacy in reinforcement learning. We analyzed the targets and impacts from the perspective of MDP, and review existing research based on the elements of MDP. Specifically, we recognize attacks and defenses of state and action, environment and reward. 

We also described the strategies and representative methods of security and privacy issues, facilitating a clear understanding of what method is used for which object of the MDP. We conducted an analysis combining those methods with the character of the elements of the MDP tuple. We went on to discuss the advantages and limitations of the studies, along with potential future directions of research into security and/or privacy in reinforcement learning. 

In summarizing the recent research into security and privacy in reinforcement learning, the following important findings can be extracted:
\begin{itemize}
\item{Every element in MDP can be attacked, and can thus affect the overall process. }
\item{Security and privacy issues in reinforcement learning are always regarded as optimal problems, and mathematical optimization methods and some intelligent learning algorithms are adopted to deal with the optimal problems. }
\end{itemize}

Based on these findings, we suggest three directions for future research:
\begin{itemize}
\item{Considering the elements in MDP. Investigating different components of MDP and the relationships between them may help to alleviate the security and privacy problems in reinforcement learning. }
\item{Considering the real environment and actual situations of the learning system is an interesting direction; for example, exploring equipment faults and environmental disturbances. }
\item{Given the many intelligent methods used in the various areas of reinforcement learning research, it would seem wise to discuss the potential problems that may occur as a result, and to improve the performance of those intelligent methods. }
\end{itemize}

\backmatter

\bmhead{Acknowledgments}

This work is supported by ARC Discovery Project (DP190100981, DP200100946) from the Australian Research Council,Australia.


\bibliography{sn-bibliography}


\end{document}